\documentclass{article}

    \PassOptionsToPackage{numbers, compress}{natbib}
\usepackage[preprint]{neurips_2026}

\usepackage[utf8]{inputenc} 
\usepackage[T1]{fontenc}    
\usepackage{hyperref}       
\hypersetup{hidelinks}
\usepackage{url}            
\usepackage{booktabs}       
\usepackage{amsfonts}       
\usepackage{nicefrac}       
\usepackage{microtype}      
\usepackage{xcolor}         

\usepackage{amsmath}
\usepackage{amssymb}
\usepackage{mathtools}
\usepackage{amsthm}

\usepackage[capitalize,noabbrev]{cleveref}

\theoremstyle{plain}
\newtheorem{theorem}{Theorem}[section]

\newtheorem{lemma}[theorem]{Lemma}

\theoremstyle{definition}

\theoremstyle{remark}

\newcommand{\nop}[1]{}
\usepackage{algorithm}
\usepackage[noend]{algorithmic}
\usepackage{amsmath,amsfonts,bm}
\usepackage{enumitem}
\usepackage{wrapfig}
\usepackage{bold-extra}
\usepackage[capitalize,noabbrev]{cleveref}
\usepackage[most]{tcolorbox}
\DeclareMathOperator*{\argmax}{arg\,max}
\DeclareMathOperator*{\argmin}{arg\,min}
\usepackage{microtype}
\usepackage{graphicx}
\usepackage{subcaption}
\usepackage{booktabs} 
\usepackage{svg}

\usepackage[disable,textsize=tiny]{todonotes}

\title{The Power of Order: Fooling LLMs with Adversarial Table Permutations}

\author{%
  Xinshuai Dong\\
  CMU \& NEC Labs\\
  \And
  Haifeng Chen\\
  NEC Labs\\
  \And
  Xuyuan Liu\\
  Dartmouth College \& NEC Labs\\
  \And
  Shengyu Chen\\
  NEC Labs\\
  \And
  Haoyu Wang\\
    NEC Labs\\
    \And
    Shaoan Xie\\
  CMU \& MBZUAI\\
  \And
  Kun Zhang\\
  CMU \& MBZUAI\\
  \And
  Zhengzhang Chen\\
      NEC Labs\\
}

\begin{document}

\maketitle

\begin{abstract}
Large Language Models (LLMs) have achieved remarkable success and are increasingly deployed in critical applications involving tabular data, such as Table Question Answering (TQA). However, 
their robustness to the structure of this input remains a critical, unaddressed question. This paper demonstrates that modern LLMs exhibit a significant vulnerability to the layout of tabular data. Specifically, 
we show that semantically-invariant permutations of rows and columns—rearrangements that do not alter the table's underlying information—are sometimes sufficient to cause incorrect or inconsistent 
model outputs. To systematically probe this vulnerability, we introduce \textbf{Adversarial Table Permutation (ATP)}, a novel, gradient-based attack that efficiently identifies worst-case permutations 
designed to maximally disrupt model performance. Our extensive experiments demonstrate that ATP significantly degrades the performance of a wide range of LLMs. This reveals a pervasive vulnerability across 
different model sizes and architectures, including the most recent and popular models. Our findings expose a fundamental weakness in how current LLMs process structured data, underscoring the urgent need to
 develop permutation-robust models for reliable, real-world applications. Code will be available. 
\end{abstract}

\section{Introduction}

Large language models (LLMs) have demonstrated powerful reasoning capabilities, leading to significant advancements in tasks involving structured data. A key area of progress is \textbf{table question answering (TQA)}, where models interpret and extract information from tables to answer natural-language questions~\citep{deng2024acl,2024naacl}. The dominant paradigm for this task involves \textbf{linearizing} the table---converting its rows and columns into a serialized text format---and including it directly in the model's prompt~\citep{zhang2024tablellm,jiang2023emnlp,ye2023large}. This approach effectively leverages the native text-processing power of LLMs, allowing them to achieve state-of-the-art performance on some TQA benchmarks without needing specialized architectural modifications.

Despite its practical effectiveness, this linearization strategy introduces a fundamental \textbf{semantic-structural mismatch}. Tables are inherently \textbf{permutation-invariant}; their underlying relational information remains unchanged regardless of the order of their rows or columns. In stark contrast, the transformer-based architectures of LLMs are fundamentally \textbf{order-sensitive} \citep{liangpearl}, processing input as a strict sequence of tokens~\citep{shi2024judging,wang2025eliminating}. This discrepancy creates a critical vulnerability. Because the model's understanding is tied to a superficial textual order, two tables containing identical information but presented in different layouts can elicit inconsistent and potentially incorrect outputs. This fragility undermines the reliability of LLMs in high-stakes applications and motivates a deeper investigation into their structural robustness.

Although previous work~\citep{yang2022acl,wang2022nips} demonstrated that row and column order can influence model predictions, its analysis has key limitations that restrict its applicability to modern systems. The methodology was primarily empirical, relying on random permutations to observe output changes without providing a systematic understanding of \textit{how} specific layouts affect model reasoning. Furthermore, this research concentrated on BERT-style models for representation learning, a paradigm fundamentally different from the now-prevalent decoder-only LLM setting, where tables are processed generatively as part of an in-context prompt. This focus on older architectures offers limited guidance on the robustness of the large-scale models used in today's applications.\looseness=-1

\begin{figure*}[!t]
 \vspace{-0.5em}
    \centering
    \includegraphics[trim={1em 10em 1em 11em}, clip, width=0.93\linewidth]{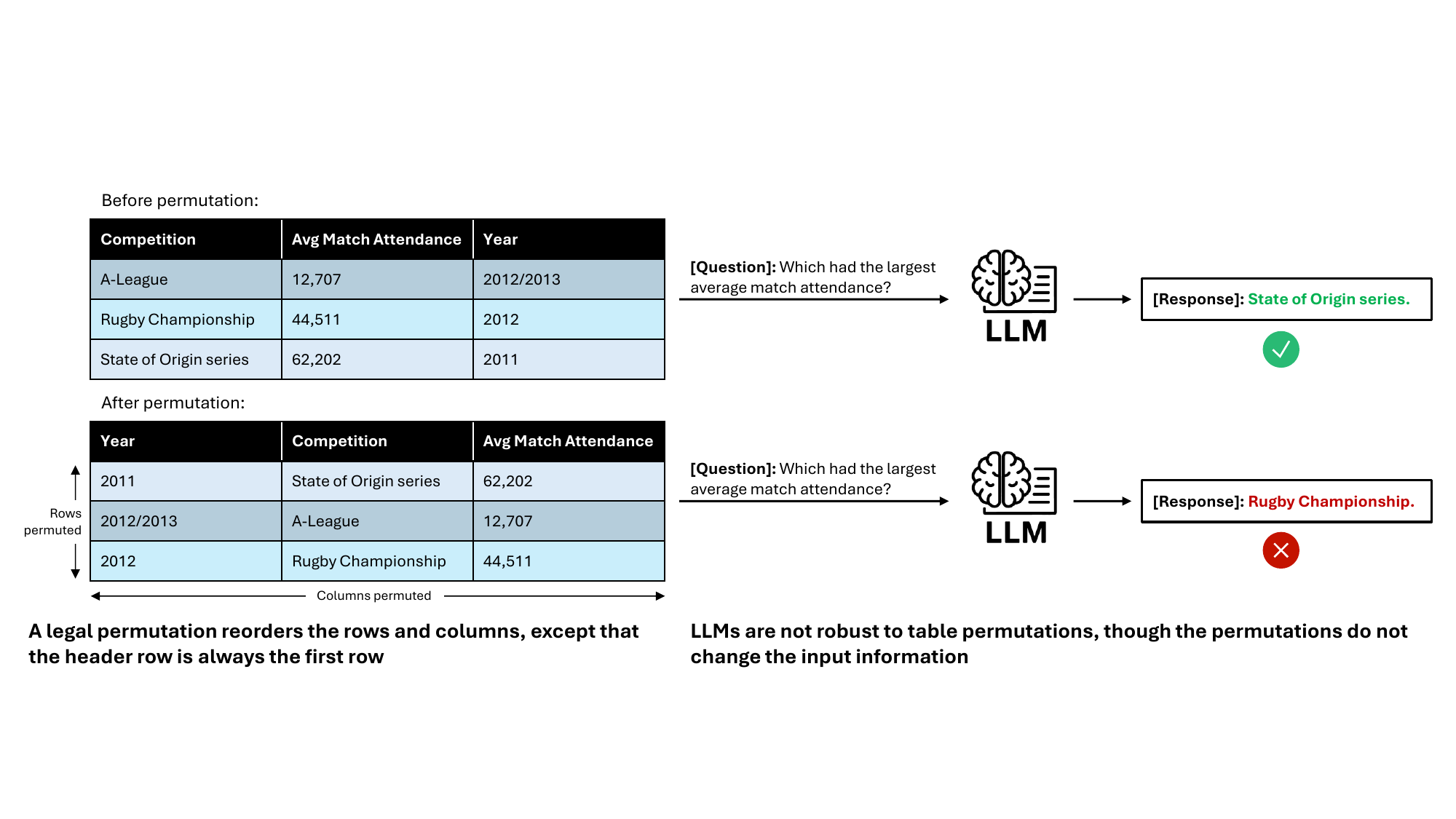}
    \caption{Illustration of the attack space for tabular inputs. A legal permutation reorders data rows and columns while keeping the header row first. Such permutations preserve the table information, but they can still fool modern LLMs.}
    \label{fig:permutation_example}
    \vspace{-1em}
\end{figure*}

In this work, we present a more rigorous investigation. We first demonstrate the susceptibility of modern LLMs to row and column permutations and then formalize this permutation sensitivity for TQA. Building on this, we introduce \textbf{Adversarial Table Permutation (ATP)}, a gradient-based attack that finds worst-case permutations by relaxing the discrete search problem into a continuous space. Across a range of instruction-tuned LLMs, ATP consistently uncovers worst-case permutations that significantly degrade prediction consistency and task performance.
An example can be found in \cref{sec:example_attack}, with mechanistic insight analysis in \cref{subsec:mechanistic_insight}. These adversarial layouts are transferable across model families and prompting styles and remain effective under commonly used prompting strategies. Our findings reveal a fundamental structural vulnerability in the prevailing ``linearize-then-prompt'' paradigm, underscoring the urgent need for 
more robust 
table reasoning techniques with LLMs.

We summarize our contributions as follows:
\nop{
\begin{itemize}[leftmargin=*, itemsep=0pt]
\item We formalize the vulnerability of current fashion LLMs against permutations on tabular data as input. More specifically,
 random row and column permutations are sometimes sufficient to fool LLMs.
\item To further highlight this vulnerability, 
We propose Adversarial Table Permutation (ATP) attack, a novel adversarial attack method that makes use of gradients to find the worst case permutation to deceive a victim model.
\item Our experiments show that ATP attack can successfully fool modern LLMs with various model sizes and types.
 This may reveal a significant weakness in current LLMs design
  and hopefully  inspire future work on LLMs towards more robust real-life tabular applications.
\end{itemize}
}

 \vspace{-0.5em}
\begin{itemize}[leftmargin=*, itemsep=0pt]
    \item We formalize the vulnerability of modern LLMs to permutations in tabular inputs, demonstrating that even random shuffling of rows and columns is sometimes sufficient to degrade model performance.

    \item To systematically expose this weakness, we propose the Adversarial Table Permutation (ATP) attack, a novel, gradient-based method that efficiently finds the worst-case permutations that cause a target model to fail.
    Our method is quite general and serves as a module that works for any open source LLM that takes embeddings, position ids, and attention masks as input.

    \item Our extensive experiments show that ATP attack successfully degrades the performance of a wide range of LLMs, and ATP consistently outperforms random and heuristic baselines.
    Additional resuls on closed LLMs also indicate that the vulnerability is not limited to small or open models.
     These findings reveal a critical design flaw in current models, underscoring the need for more robust architecture for real-world tabular applications.

\end{itemize}

\vspace{-1em}
\section{Problem Setting}
\label{sec:problem_setting}
\vspace{-0.2em}

\subsection{Attack Space for Tabular Input: Row and Column Permutations}
\label{subsec:attack_space}
\vspace{-0.2em}


Given a table and an order-insensitive question, a robust TQA model should produce an answer that is invariant to row and column permutations that preserve the table's meaning. By \emph{order-insensitive},
 we mean that the question does not explicitly depend on presentation order, such as asking for the content of the first row or last column. Such order-sensitive questions form only a small fraction of the benchmarks we study (around 1\% on average).
  We filter these examples out and focus on the TQA setting in which table permutations should not change the correct answer.
 Specifically, 
given a table with $n+1$ rows and $m$ columns where the first is a {header row}, 
one can arbitrarily permute the remaining $n$ data rows and all $m$ columns without changing the underlying relational information. An example of such a semantically equivalent permutation can be seen in \cref{fig:permutation_example},
where the original and permuted tables contain the same information.

Formally, we define this attack space in the context of Table Question Answering (TQA) tasks, where we have i.i.d. samples of an input table $\mathbf{T}$, a question $\mathbf{Q}$, and an answer $\mathbf{A}$ from a given data distribution. Here, $\mathbf{Q}$ and $\mathbf{A}$ are both sequences of words, while $\mathbf{T}$ is represented as an $(n+1) \times m$ matrix where each cell contains a sequence of words. 
 %
 Let $\Pi_{k}$ be the set of all $k \times k$ permutation matrices. Then the attack space for the input table $\mathbf{T}$ is defined as:
 \begin{align}
    \label{eq:attackspace}
 \bm{P}_r \mathbf{T} \bm{P}_c, ~\text{s.t.,~} \bm{P}_r \in \hat{\Pi}_{n+1}, \bm{P}_c \in\Pi_{m},~
 \text{where~} \hat{\Pi}_{n+1}:= \{\bm{P}\in\Pi_{n+1}: \bm{P}_{[0,0]=1}\}.
 \end{align}

In \cref{eq:attackspace}, the matrix $\bm{P}_r$ permutes the rows of $\mathbf{T}$ and $\bm{P}_c$ permutes the columns. The constraint $\bm{P}_{[0,0]}=1$ in the definition of $\hat{\Pi}_{n+1}$ ensures that the header row is always the first row. Given this formulation, we next discuss the key research problems this work aims to address.

\vspace{-0.5em}
\subsection{Are LLMs Robust Against Table Permutations?}
\vspace{-0.2em}
The key motivation of this work is to investigate to what extent current LLMs are robust against 
row and column permutations of the input table.
This can be decomposed into three key research questions:
(i) Are current LLMs robust to random table permutations?
(ii) How to generate the worst-case table permutation to fool a LLM?
(iii) To what extent current LLMs are robust to the worst-case table permutations?
We will first formalize (i) and (ii) in the rest of this section and then 
 address (ii) in \cref{sec:method} by proposing a novel attack method,
and finally answer (i) and (iii) in our experiments in \cref{sec:exp}.

Consider  a LLM that parametrizes a probability mass function
 over the natural language space, as $P_{\rm{model}}(\cdot)$.
 To employ the LLM for TQA tasks, 
 we generate the model response by sampling from the parameterized conditional distribution,
 as  $\hat{\mathbf{A}} \sim P_{\rm{model}}(\cdot|~\mathbf{T},\mathbf{Q})$.
 Then we evaluate to what extent $\hat{\mathbf{A}}$ is semantically aligned with the ground truth $\mathbf{A}$,
 by some evaluation metrics $\mathcal{M}(\mathbf{A}, \hat{\mathbf{A}})$ (the bigger the better alignment).

Thus, the research problem (i) can be formulated as
the calculation of the following
 \begin{align}
 \label{eq:random_attack}
 \mathbb{E}_{ \hat{\mathbf{A}}  \sim P_{\rm{model}}(\cdot|~\bm{P}_r \mathbf{T} \bm{P}_c, \mathbf{Q}),~ 
 \bm{P}_r \sim \mathcal{U}_r,~
 \bm{P}_c \sim \mathcal{U}_c
 } ~\mathcal{M}(\mathbf{A}, \hat{\mathbf{A}}),
 \end{align}
where $\mathcal{U}_r$ and 
 $\mathcal{U}_c$ are uniform distribution over $\hat{\Pi}_{n+1}$
 and  $\Pi_{m}$, respectively.

 As for (ii), it can be formulated as finding the worst case combination of row and column permutations $(\bm{P}_r^*,\bm{P}_c^*)$ to fool a victim model $P_{\rm{model}}$, as,
\begin{align}
\label{eq:ori_worst_case_permutation}
 (\bm{P}_r^*,\bm{P}_c^*) =  
 \argmin_{ \bm{P}_r \in\hat{\Pi}_{n+1},\bm{P}_c \in {\Pi}_{m}} ~ \mathbb{E}_{ \hat{\mathbf{A}}
   \sim P_{\rm{model}}(\cdot|~\bm{P}_r \mathbf{T} \bm{P}_c, \mathbf{Q})} ~\mathcal{M}(\mathbf{A}, \hat{\mathbf{A}}),
 \end{align}
 and then evaluate the performance under such worst case permutations, as
 \begin{align}
 \mathbb{E}_{ \hat{\mathbf{A}}^*
   \sim P_{\rm{model}}(\cdot|~\bm{P}_r^* \mathbf{T} \bm{P}_c^*, \mathbf{Q})} ~\mathcal{M}(\mathbf{A}, \hat{\mathbf{A}}^*).
 \end{align}

 By \cref{eq:random_attack}, it is straightforward to evaluate the robustness of a LLM against random table permutations.
 As a contrast, the optimization problem in \cref{eq:ori_worst_case_permutation} is highly nontrivial. Solving the  combinatorial
optimization problem in \cref{eq:ori_worst_case_permutation}  directly in the
  permutation space $\hat{\Pi}_{n+1}$ and ${\Pi}_{m}$ is NP-hard. The computation complexity grows exponentially with the shape of the input table. For example, when $n=m=8$, there are around $1.6\times 10^{9}$ different kinds of combinations of row and column permutations, and this number increases to $1.3\times 10^{11}$ when $n$ and $m$ are increased  by only 1.
Therefore, it is crucial to have a more effective way to find the worst case permutation, and we  propose our novel method in what follows.

 \begin{figure*}[t]
    \centering
    \vspace{-0.5em}
    \includegraphics[trim={1em 10em 1em 6em}, clip, width=0.9\linewidth]{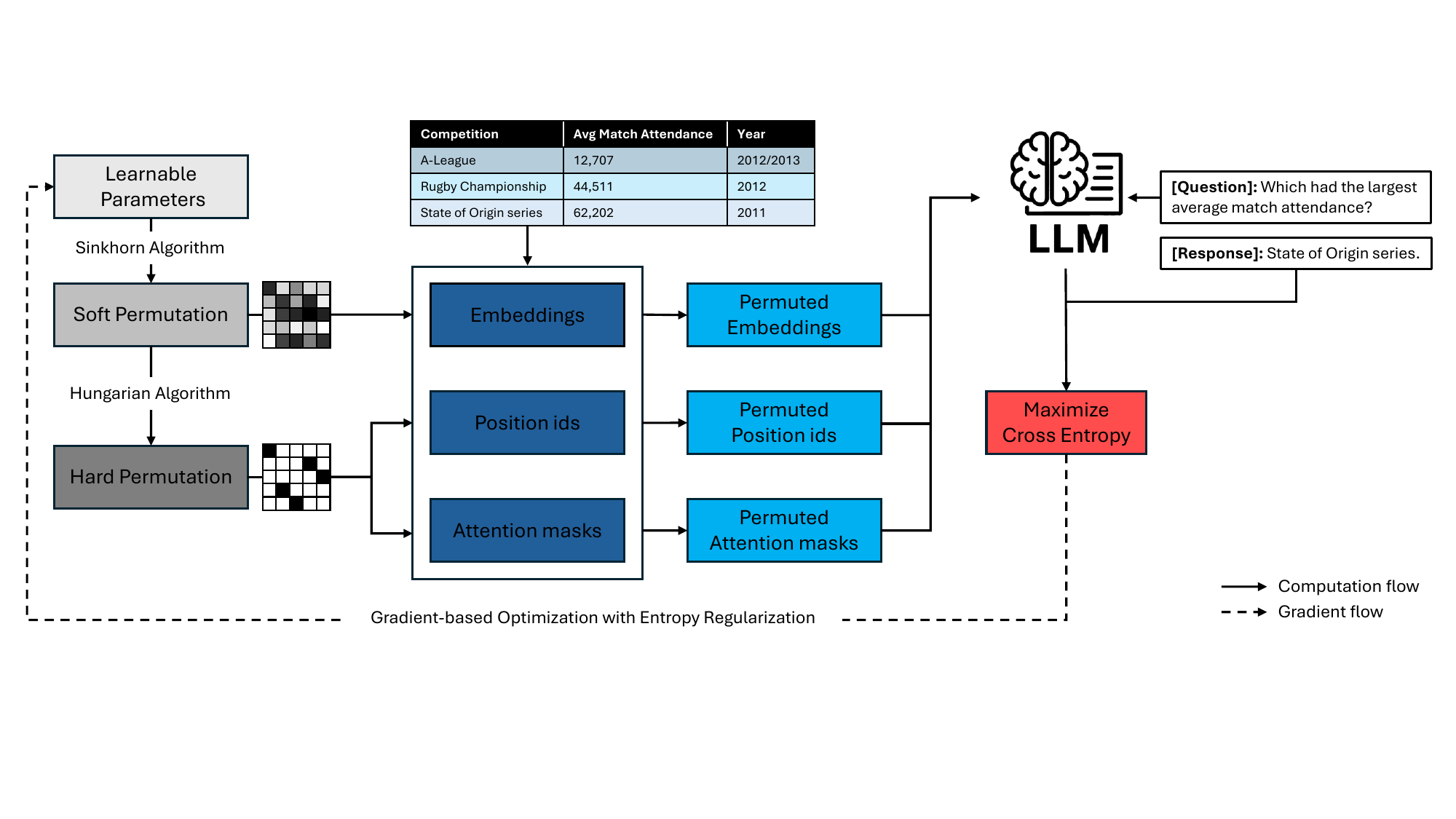}
    \vspace{-1em}
    \caption{Overview of ATP. Learnable parameters define soft permutations (doubly stochastic matrices), which are applied to table embeddings. Their hard projections are applied to position IDs and attention masks. The permuted input, question, and reference answer are fed to the victim LLM, and gradients update the permutation parameters to maximize the attack objective.}
    \vspace{-1em}
    \label{fig:procedure}
\end{figure*}

\vspace{-0.5em}
\section{Adversarial Table Permutation Attack}
\label{sec:method}
\vspace{-0.2em}

The core challenge in finding the worst-case permutation, as formulated in \cref{eq:ori_worst_case_permutation}, is that it requires optimizing over a vast and discrete space of permutation matrices, which is computationally intractable for tables of non-trivial size.
To overcome this hurdle, our proposed Adversarial Table Permutation (ATP) attack reframes this
discrete problem into a continuous one that can be solved efficiently with gradient-based methods.
This is achieved through two key relaxations, which are detailed below. An illustration of the overall process can be found in \cref{fig:procedure}.

\vspace{-0.5em}
\subsection{Relaxing the Discrete Problem into a Differentiable One}
\label{sec:relaxation}
\vspace{-0.2em}

\paragraph{From a Non-Differentiable Metric to a Differentiable Loss.}
Our first step is to transform the optimization objective into a continuous, differentiable form.
Specifically,
the evaluation metric $\mathcal{M}(\mathbf{A}, \hat{\mathbf{A}})$ is often non-differentiable 
and requires sampling model outputs $\hat{\mathbf{A}}$, leading to noisy and unstable gradients. We replace this objective with the maximization of the  cross-entropy loss $\mathcal{L}_{\text{CE}}$ of the ground-truth answer $\mathbf{A}$. This provides a differentiable proxy that directly measures the alignment of model output with the ground truth. The optimization problem thus becomes
\begin{align}
\label{eq:worst_case_permutation_a}
 (\bm{P}_r^*,\bm{P}_c^*) =  
 \argmax_{ 
  \bm{P}_r \in\hat{\Pi}_{n+1},\bm{P}_c \in {\Pi}_{m}
 }~ \mathcal{L}_{\text{CE}}(P_{\rm{model}},\bm{P}_r \mathbf{T} \bm{P}_c, \mathbf{Q},\mathbf{A}),
\end{align}
where the cross-entropy loss is as follows:
\begin{align}
 \label{eq:def_ce}
 \mathcal{L}_{\text{CE}}(\cdot) = - \sum_{t=0}^{|\mathbf{A}|-1} \log P_{\rm{model}} ( \mathbf{A}_{[t]} | \bm{P}_r \mathbf{T} \bm{P}_c, \mathbf{Q}, \mathbf{A}_{[:t]}).
\end{align}
Here $\mathcal{L}_{\text{CE}}$ is the sum of the cross-entropy for each token $\mathbf{A}_{[t]}$ conditioned on the correct context. It provides a stable, gradient-friendly objective without the need for sampling.

\vspace{-0.5em}
\paragraph{From Permutation Matrices to Doubly Stochastic Matrices.}
While the objective is now differentiable, the search space of permutation matrices ($\hat{\Pi}_{n+1}$ and $\Pi_{m}$) remains discrete. To create a continuous search space, we perform a convex relaxation to relax the set of permutation matrices to its convex hull.
The reasons to consider convex hull lies in that the convex hull is the minimal convex and continuous superset of the original discrete space, which facilitates efficient optimization. 

\begin{lemma} [Birkhoff-von Neumann \citep{birkhoff1946tres}]
\label{thm:birkhoff}
The convex hull of the set of $n\times n$ permutation matrices, $\Pi_{n}$, is the set of $n\times n$ doubly stochastic matrices, $\mathbb{D}_{n}$.
\end{lemma}

By Lemma \ref{thm:birkhoff}, 
the convex hull of ${\Pi}_m$ is the set of all $m\times m$ doubly stochastic matrices, $\mathbb{D}_{m}$.
As for relaxing $\hat{\Pi}_{n+1}$,
we further define $\hat{\mathbb{D}}_{n+1}=\{\bm{D}\in\mathbb{D}_{n+1}: \bm{D}_{[0,0]}=1\}$,
the set of all $n\times n$ doubly stochastic matrices whose upper-left entry is always 1.
As such we can now optimize over the continuous and convex sets of doubly stochastic matrices $\hat{\mathbb{D}}_{n+1}$ and $\mathbb{D}_{m}$.

Specifically, we first parametrize two  unconstrained real matrices $\bm{\theta}_r \in \mathbb{R}^{n\times n}$ and $\bm{\theta}_c \in \mathbb{R}^{m\times m}$,
and then transform they to two "soft" permutation matrices $\bm{D}_r$ and $\bm{D}_c$,
by leveraging the differentiable Sinkhorn algorithm \citep{sinkhorn1964relationship,adams2011ranking,mena2018learning}, as
\vspace{-0.2em}
\begin{align}
\label{eq:sinkhorn}
\bm{S}^0 (\bm{\theta}) = \exp(\bm{\theta}),~
\bm{S}^{i+1} (\bm{\theta}) = \bm{N}_c(\bm{N}_r(\bm{S}^i (\bm{\theta}))),~
\bm{S} (\bm{\theta}) =  \lim_{i\to \infty} \bm{S}^{i} (\bm{\theta}),
 \end{align}
 where $\bm{N}_r$ and $\bm{N}_c$ are row normalization and column normalization, as: 
  \vspace{-0.2em}
\begin{align}
\label{eq:row_col_normalize}
\bm{N}_r(\bm{X})_{ij} = \frac{\bm{X}_{ij}}{\sum_{j'} \bm{X}_{ij'}},~
\bm{N}_c(\bm{X})_{ij} = \frac{\bm{X}_{ij}}{\sum_{i'} \bm{X}_{i'j}}.
 \end{align}
By the theorem in \cite{sinkhorn1964relationship}, 
we have $\bm{S}(\bm{\theta}_r) \in \mathbb{D}_{n}$
and $\bm{S}(\bm{\theta}_c) \in \mathbb{D}_{m}$,
and then we define the soft permutation matrices $\bm{D}_r$  and $\bm{D}_c$, as  
\vspace{-0.2em}
\begin{align}
\label{eq:calculate_drdc}
{\bm{D}_{r}}_{[1:,1:]}=\bm{S}(\bm{\theta}_r),~
{\bm{D}_{r}}_{[0,0]}= 1,~
{\bm{D}_{r}}_{[0,1:]}={\bm{D}_{r}}_{[1:,0]}=0, 
\text{and} ~ \bm{D}_{c}=\bm{S}(\bm{\theta}_c),
\vspace{-0.5em}
\end{align}
where ${\bm{D}_{r}}$ is designed to ensure the header row remains fixed,
and thus we have ${\bm{D}_{r}}\in\hat{\mathbb{D}}_{n+1}$
and ${\bm{D}_{c}}\in {\mathbb{D}}_{m}$.
This allows us to search for the optimal permutation in a continuous space using gradients with respect to $\bm{\theta}_r$ and $\bm{\theta}_c$.

\vspace{-0.5em}
\subsection{Projecting Back to the Permutation Space}
\label{sec:projection}
\vspace{-0.2em}

The relaxation in \cref{sec:relaxation} allows for gradient-based optimization, but it introduces a new challenge: the optimized matrices $\bm{D}_r$ and $\bm{D}_c$ are  ``soft'' permutations, not the hard, discrete permutations required for a valid attack. Furthermore, an LLM's input for table $\mathbf{T}$ consists of multiple components: continuous token embeddings $\mathbf{T}^{\text{emb}}$ and discrete position ids $\mathbf{T}^{\text{pos}}$ and attention masks $\mathbf{T}^{\text{att}}$. While soft permutations can be applied to continuous embeddings, they cannot be applied to discrete inputs.

To address this, we apply the permutations differently based on the input type. We use the soft doubly stochastic matrices for the embeddings and project them back to the nearest permutation matrices for the discrete components. Such a projection is captured by a maximum weight matching problem that can be solved by the Hungarian algorithm \citep{kuhn1955hungarian,kuhn1956variants}, as
\vspace{-0.2em}
\begin{align}
 \text{Proj}_n(\bm{D}) = \argmax_{\bm{P}\in\Pi_{n}} {\langle \bm{P}, \bm{D} \rangle}_{F}, \bm{D}\in \mathbb{D}_n,
\end{align}
where $\langle \cdot, \cdot \rangle_F$ is the Frobenius inner product and the subscript $n$ in $\text{Proj}_n(\cdot)$ is dropped when the context is clear.
The input table to the model is thus permuted in a hybrid mode, as
\vspace{-0.2em}
\begin{align}
\label{eq:permute_diff}
P_{\rm{model}}(\cdot| ~
\bm{D}_r \mathbf{T}^{\text{emb}} \bm{D}_c,~
\text{Proj}(\bm{D}_r) \mathbf{T}^{\text{pos}} \text{Proj}(\bm{D}_c),~ 
\text{Proj}(\bm{D}_r) \mathbf{T}^{\text{att}} \text{Proj}(\bm{D}_c),~
\mathbf{Q}).
\end{align}
In Eq. \ref{eq:permute_diff} we use soft permutations to permute the embeddings and the hard permutations to permute positional ids and attention masks, as illustrated in \cref{fig:procedure}.
This hybrid approach allows us to maintain a differentiable optimization pipeline while ensuring the final generated attack is valid and correctly manipulates all aspects of the model's input.

\vspace{-0.5em}
\subsection{Regularization: Information Entropy-based Over Temperature-based}
\label{sec:regularization}
\vspace{-0.2em}

Another key challenge in our relaxed optimization is to ensure that the resulting doubly stochastic matrices, $\bm{D}_r$ and $\bm{D}_c$, are close to actual permutation matrices. Without this constraint, the soft permutations during optimization could converge to solutions far from any single permutation (e.g., a uniform matrix), creating a significant gap between the loss measured during optimization and the attack's true effectiveness.
Thus, we must encourage the soft matrices to be ``sharp'' and structurally similar to a hard permutation.

 A classic strategy is to follow Gumbel-Softmax \citep{jang2016categorical,maddison2016concrete}
to introduce temperature $\tau$ into the Sinkhorn algorithm, as $\bm{S}(\bm{\theta}/\tau)$. 
Lowering $\tau$ is analogous to pushing the optimization towards low-entropy  solutions and thus closer to a hard permutation.
However, in practice, we found that to encourage $\bm{S}(\bm{\theta}/\tau)$ to be sufficiently close to a hard permutation, it relies on a pretty small temperature $\tau$ (e.g., $\tau \leq 0.05$); such a small $\tau$ introduces significant computational instability. This issue is exacerbated when using low-precision floating-point formats like float16 or bfloat16, which are however very common for LLMs, especially in memory-constrained scenarios.

To circumvent this instability while still achieving the same goal, we incorporate the information entropy $\mathcal{H}(\cdot)$ as an explicit regularization term in our final objective, as follows:
\vspace{-0.5em}
\begin{align}
\label{eq:entropy}
\mathcal{H}(\bm{D}) = -{\sum}_{i=1}^n {\sum}_{j=1}^n \bm{D}_{ij} \log(\bm{D}_{ij}),
\end{align}
where the input $\bm{D}$ is a $n\times n$ doubly stochastic matrix.
This approach, partly inspired by \cite{dong2021towards}, directly encourages the soft matrices to be close to permutation matrices without the numerical issues associated with a small temperature $\tau$. We also empirically validate this point by our ablation study in \cref{sec:ablation}.

\vspace{-0.5em}
\subsection{Final Objective}
\vspace{-0.2em}

Thus, by adding the entropy term to the optimization objective, 
combining the differentiable loss and the hybrid permutation strategy, our final optimization objective is to find the parameters $(\bm{\theta}_r^*, \bm{\theta}_c^*)$ that maximizes the weighted sum of cross-entropy loss and entropy regularization terms:
\vspace{-0.5em}
\begin{align}
\label{eq:final_without_regularization}
\vspace{-0.5em}
 (\bm{\theta}_r^*,\bm{\theta}_c^*) = 
 \argmax_{\bm{\theta}_r \in \mathbb{R}^{n\times n},\bm{\theta}_c \in \mathbb{R}^{m\times m}} 
&\mathcal{L}_{\text{CE}}(  
     P_{\rm{model}},~
\bm{D}_r \mathbf{T}^{\text{emb}} \bm{D}_c,
    \text{Proj}(\bm{D}_r) \mathbf{T}^{\text{pos}} \text{Proj}(\bm{D}_c),  \nonumber \\
    &\text{Proj}(\bm{D}_r) \mathbf{T}^{\text{att}} \text{Proj}(\bm{D}_c),
    \mathbf{Q},\bm{A}) -\lambda_1\mathcal{H}(\bm{D}_r)  - \lambda_2\mathcal{H}(\bm{D}_c),
    \vspace{-0.5em}
 \end{align}
where $\lambda_1$ and $\lambda_2$ are hyper-parameters controlling the weight of the entropy terms.
We note that $\mathcal{L}_{\text{CE}}$ in \cref{eq:final_without_regularization} can also be replaced by a KL divergence loss to maximize the discrepancy between the clean generation and the permuted generation,
which does not require label / ground truth response.
In our experiments \cref{sec:exp}, we mainly focus on the target attack with $\mathcal{L}_{\text{CE}}$, while we also reports the attack performance using KL loss in \cref{subsec:kl_attack_result} to show the extensibility of ATP to the label free setting.

Once the solution $(\bm{\theta}_r^*, \bm{\theta}_c^*)$ are found,
the resulting optimal soft matrices $(\bm{D}_r^*, \bm{D}_c^*)$ are projected via $\text{Proj}(\cdot)$ to obtain the final adversarial permutation matrices $(\bm{P}_r^*, \bm{P}_c^*)$ used to attack the LLM.
We solve the optimization in \cref{eq:final_without_regularization}  by Adam \citep{kingma2014adam}, where the number of iterations $N_{\text{attack}}$ serves as a hyper-parameter (ablation study on $N_{\text{attack}}$ in \cref{sec:ablation}).
We also summarize the whole algorithm procedure of the proposed ATP attack in \cref{alg:ATP}.

\begin{algorithm}[t]
   \caption{Adversarial Table Permutation (ATP) Attack}
   \label{alg:ATP}
   \begin{algorithmic}[1]
    \STATE {\bfseries Input:~} $P_{\rm{model}}$, $\mathbf{T},\mathbf{Q},\mathbf{A}$, $N_{\text{attack}}$, $\lambda_1$, $\lambda_2$;
  \STATE {\bfseries Output:~} Worst-case row permutation $\bm{P}_r^*$ and column permutation $\bm{P}_c^*$;
    \STATE Initialize $\bm{\theta}_r,\bm{\theta}_c$ and get $\mathbf{T}^{\text{emb}},\mathbf{T}^{\text{pos}},\mathbf{T}^{\text{att}}$;\;
  \FOR{$i=1$ to $N_{\text{attack}}$}
      \STATE Calculate $\bm{S}(\bm{\theta}_r)$ and $\bm{S}(\bm{\theta}_c)$ by \cref{eq:sinkhorn}; \;
    \STATE Calculate $\bm{D}_r$ and $\bm{D}_c$ by \cref{eq:calculate_drdc}
    and $\mathcal{H}(\bm{D}_r)$ and $\mathcal{H}(\bm{D}_c)$ by \cref{eq:entropy}
    ; \;
    \STATE Calculate $\text{Proj}(\bm{D}_r)$ and $\text{Proj}(\bm{D}_c)$ by the Hungarian algorithm \citep{kuhn1955hungarian};\;
     \STATE Calculate $\mathcal{L}_{\text{CE}}$ in \cref{eq:final_without_regularization};\;
       \STATE Calculate the gradient of $\mathcal{L}_{\text{CE}}$ and update 
       $\bm{\theta}_r$ and $\bm{\theta}_c$
       by Adam \citep{kingma2014adam} to get $\bm{\theta}_r^*,\bm{\theta}_c^*$;\;
  \ENDFOR
    \STATE{Calculate $\bm{D}_r^*,\bm{D}_c^*$ given  
    $\bm{\theta}_r^*,\bm{\theta}_c^*$ by \cref{eq:calculate_drdc};\;
    \STATE Calculate $\bm{P}_r^*=\text{Proj}(\bm{D}_r^*)$ and $\bm{P}_c^*=\text{Proj}(\bm{D}_c^*)$}\;
    \STATE {\bfseries return~$\bm{P}_r^*$, $\bm{P}_c^*$}
   \end{algorithmic}
   \end{algorithm}

\vspace{-0.5em}
\section{Related Work}
\vspace{-0.5em}

\paragraph{Adversarial attacks and robustness for LLMs.}
LLMs remain vulnerable to adversarial manipulations that bypass safety mechanisms or induce undesired behavior~\citep{he2024your,qi2024iclr,hsiung2025llm}. 
Token- and sentence-level perturbations can reliably elicit harmful outputs~\citep{dong2021towards,dong2021should,zhao2022certified,ye2022aaai,huang2021eacl},
 and audits of ChatGPT have revealed substantial robustness weaknesses~\citep{wang2023robustness}. Beyond input-level perturbations, black-box jailbreak frameworks automate the discovery of exploit
  templates~\citep{yu2023gptfuzzer}. A small number of harmful instruction-response exemplars can also act as few-shot triggers that compromise alignment~\citep{qi2024iclr}, 
  and persuasion-based prompt generation has been used to produce effective jailbreaks~\citep{2024acl}. These findings motivate principled robustness analyses for LLMs beyond conventional text perturbations.

\looseness=-1

\vspace{-0.5em}
\paragraph{Comparison with PEARL \citep{liangpearl}.}
Our work is also related to PEARL, which reorders options / examples in prompt to attack LLMs, but ATP differs significantly in both attack space and optimization methodology. 
First,  while PEARL focuses on one-dimensional permutations of options or examples within a prompt, ATP operates on the two-dimensional structure of tabular data, necessitating a more complex permutation space that keeps the header row fixed while reordering all other rows and columns. 
Second, the optimization strategies diverge in stability and execution; whereas PEARL utilizes the Sinkhorn operator with a temperature parameter and the Gumbel-max trick to sample hard permutations, ATP avoids the computational instability of low-temperature Sinkhorn operations by using information entropy regularization. This approach forces soft permutations to converge toward hard ones without sacrificing numerical stability, especially in the low-precision formats common to LLMs. 
Lastly, instead of stochastic sampling, ATP employs the Hungarian algorithm to project soft permutations onto a discrete space. To maintain end-to-end differentiability while ensuring validity, ATP uses a hybrid application: soft permutations are applied to continuous token embeddings, while hard permutations are used for discrete positional IDs and attention masks. This makes ATP a more stable and targeted tool for exposing the structural vulnerabilities of LLMs in tabular data reasoning.
\looseness=-1

\vspace{-0.5em}
\paragraph{LLMs for TQA.}
Table-based reasoning has advanced with LLMs and their emergent reasoning capabilities~\citep{deng2024acl,2024naacl,su2024tablegpt2,wang2024chain}. A common practice is to linearize table contents into text and include them in the prompt. \cite{rajkumar2022evaluating} use in-context examples for SQL generation, while \cite{cheng2023iclr} prompt LLMs to generate executable programs via SQL APIs. \cite{lin-etal-2023-inner} extract sub-tables containing relevant information, \cite{ye2023sigir} improve end-to-end reasoning by decomposing table contexts and questions, and \cite{nguyen2025tmlr} decompose queries into atomic steps for interpretable answers.
 
Recent work also questions whether serialized text is an adequate interface for tables. \cite{wang2026beyond} propose TabGR, a training-free framework that converts a table into an attributed table graph, uses question-guided Personalized PageRank to surface relevant triples, and enables graph-grounded reasoning paths. \cite{li2025table} propose TaMo, which treats tables as an independent modality by aligning a hypergraph-enhanced table encoder with LLM token embeddings. These approaches seek more structure-aware interfaces. In contrast, our goal is diagnostic: we expose a failure mode of existing linearized LLM inference by optimizing semantics-preserving row and column permutations.
\looseness=-1

\vspace{-0.5em}
\paragraph{Synthetic tabular data.}
A separate line of work studies how to generate realistic synthetic tabular data. Classical neural synthesizers adapt generative models to heterogeneous columns: CTGAN uses a conditional GAN with mode-specific normalization and training-by-sampling for mixed continuous and discrete features~\citep{xu2019modeling}, while TabDDPM applies diffusion modeling to tabular data and supports both continuous and categorical attributes~\citep{kotelnikov2023tabddpm}. More recent work uses language models as tabular data generators. GReaT serializes table rows as text and fine-tunes an autoregressive language model to sample realistic records~\citep{borisov2022language}. REaLTabFormer extends this idea to non-relational and relational tables using GPT-2 for single-table synthesis and a sequence-to-sequence model conditioned on parent rows for relational data~\citep{solatorio2023realtabformer}. Tabby modifies the Transformer architecture with column-aware gated mixture-of-experts for structured data synthesis~\citep{cromp2025tabby}. These works generate tabular data distributions, whereas we study whether LLMs reason robustly over a fixed table under semantically equivalent orderings.
 \looseness=-1
 
\vspace{-0.5em}
\paragraph{Robustness Evaluation for TQA.} 
Several studies have highlighted the robustness limitations of TQA systems. For instance, \cite{chen2023nips} introduces permutation-invariant table representations, while \cite{wang2022nips} extends this approach to multi-table scenarios. \cite{bhandari2025tmlr} investigates the robustness of LLMs for TQA under domain shift, and \cite{yang2022acl} demonstrates that row and column order can significantly influence model predictions. In related work, \cite{zong2023fool} prompt LLMs to generate adversarial examples to improve model robustness during training. However, these investigations are largely empirical—typically involving random permutations of table structures followed by performance observation—thus failing to systematically characterize the worst-case effects of structural perturbations under in-context LLM inference.
\looseness=-1

\begin{table*}[t]
\vspace{-1em}
  \centering
  \caption{LLM-as-judge alignment scores for open LLMs on the WTQ evaluation set under different attacks. Lower scores indicate worse response alignment and stronger attacks.}
  \vspace{-0.5em}
  \label{tab:wtq_result}
   \setlength{\tabcolsep}{1.0pt}
  \renewcommand{\arraystretch}{1.05}
  \begin{tabular}{l| c|cc|ccc|ccc}
    \toprule
    \multicolumn{1}{c|}{} & \multicolumn{9}{c}{\textbf{WTQ Dataset Evaluation Set}} \\
    \cmidrule(lr){2-10}
    \multicolumn{1}{l|}{\textbf{LLMs}} &  & \multicolumn{2}{c|}{\textbf{Random Perm}} & \multicolumn{3}{c|}{\textbf{Heuristics}} & \multicolumn{3}{c}{\textbf{ATP Attack}}  \\
    \cmidrule(lr){2-2}\cmidrule(lr){3-4}\cmidrule(lr){5-7}\cmidrule(lr){8-10}
       & \textbf{Vanilla} & \textbf{Rand} &  \textbf{Best 20} & \textbf{Row Rvs} & \textbf{Col Rvs} & \textbf{E-Search}  &\textbf{Row} & \textbf{Col}  & \textbf{Row\&Col}  \\
    \midrule
    \textsc{Llama-3.1-8B}               & 0.26 & 0.17 & 0.16 & 0.25 & 0.21 & 0.16 & 0.22 & 0.18 & \textbf{0.13} \\
    \textsc{Llama-3.1-8B-Inst}           & 0.46 & 0.31 & 0.29 & 0.41 & 0.38 & 0.28 & 0.35 & 0.31 & \textbf{0.22} \\
    \textsc{TableLLM-8B}                & 0.33 & 0.23 & 0.19 & 0.31 & 0.26 & 0.18 & 0.28 & 0.24 & \textbf{0.16} \\
    \textsc{Qwen2.5-1.5B-Inst}            & 0.14 & 0.08 & 0.06 & 0.10 & 0.09 & 0.08 & 0.10 & 0.09 & \textbf{0.04} \\
    \textsc{Qwen2.5-3B-Inst}            & 0.26 & 0.19 & 0.14 & 0.21 & 0.21 & 0.14 & 0.20 & 0.17 &  \textbf{0.11} \\
    \textsc{Qwen2.5-7B-Inst}            & 0.28 & 0.21 & 0.18 & 0.27 & 0.24 & 0.17 & 0.25 & 0.22 &  \textbf{0.12} \\
    \textsc{Qwen2.5-14B-Inst}            & 0.47 & 0.33 &0.30 & 0.42 & 0.38 & 0.29 & 0.38 & 0.36 & \textbf{0.26} \\
    \textsc{CodeLlama-7B-Inst}           & 0.18 & 0.16 & 0.14 & 0.17 & 0.14 & 0.11 & 0.16 & 0.14 & \textbf{0.09} \\
    \textsc{DSR1-Dt-Llama-8B}        & 0.24 & 0.15 & 0.14 & 0.22 & 0.18 & 0.16 & 0.19 & 0.16 & \textbf{0.09} \\
    \textsc{DSR1-Dt-Qwen-7B}          & 0.15 & 0.12 & 0.11 & 0.13 & 0.13 & 0.09 & 0.12 & 0.11 & \textbf{0.06} \\
    \bottomrule
  \end{tabular}
  \vspace{-0.5em}
\end{table*}

\vspace{-1em}
\section{Experiments}
\label{sec:exp}
\vspace{-0.5em}

This section answers the robustness questions posed in \cref{sec:problem_setting}: How sensitive are current LLMs to random table permutations, 
and how much worse are optimized adversarial permutations? We also report ablations that isolate the contribution of ATP's design choices, gain mechanistic insight on the vulnerability, and summarize runtime costs.

\vspace{-1em}
\subsection{Experimental Settings}
\vspace{-0.2em}

\textbf{Datasets.} We evaluate on three widely used document-embedded TQA datasets: WTQ~\citep{pasupat2015compositional}, TATQA~\citep{zhu2021tat}, and FeTaQA~\citep{nan2022fetaqa}, following \cite{zhang2024tablellm}. For each dataset, we report vanilla and attacked performance on the evaluation set after filtering order-sensitive questions. The average table sizes are 10 rows by 5 columns for WTQ, 10 rows by 4 columns for TATQA, and 12 rows by 5 columns for FeTaQA. The average linearized input lengths are approximately 145, 140, and 160 whitespace tokens, respectively. Numeric questions account for 38\% of WTQ and 43\% of TATQA, while lookup questions account for 68\% of FeTaQA.
 
Some benchmark questions explicitly depend on presentation order, such as questions asking about the first or last row. Such examples should not be invariant to table permutations and are outside our scope. We identify candidate order-sensitive questions using keywords such as ``first'', ``last'', and ``top'', followed by human verification. They account for 3.0\% of WTQ, 0\% of TATQA, and 0.6\% of FeTaQA, and we remove them from the experiments.
 
\textbf{Evaluation metric.} We evaluate response quality with an LLM-as-judge score, following \cite{zheng2023nips}. In TQA, lexical-overlap metrics such as exact match, BLEU~\citep{papineni2002bleu}, and ROUGE-L~\citep{lin2004rouge} can misrepresent semantic correctness: answers may be lexically different but equivalent, or lexically similar but wrong. We therefore use Gemini-2.5~\citep{comanici2025gemini} as a held-out judge to score alignment between the reference answer and the model response on a 0--1 scale. The table itself is not shown to the judge, so the judge is not directly exposed to the table-order perturbation. A human-evaluation study over 500 examples shows that the LLM-as-judge score has the highest Spearman rank correlation with human ratings (about 0.85), exceeding ROUGE-L by a clear margin. Details are provided in \cref{sec:llm_as_judge,sec:compare_metric}.
\looseness=-1

\begin{table*}[t]
\vspace{-0em}
  \centering
 \caption{LLM-as-judge alignment scores for open LLMs on the TATQA evaluation set under different attacks. Lower scores indicate worse response alignment and stronger attacks.}
  \vspace{-0.5em}
  \label{tab:tatqa_result}
   \setlength{\tabcolsep}{1.0pt}
  \renewcommand{\arraystretch}{1.05}
  \begin{tabular}{l| c|cc|ccc|ccc}
    \toprule
    \multicolumn{1}{c|}{} & \multicolumn{9}{c}{\textbf{TATQA Dataset Evaluation Set}} \\
    \cmidrule(lr){2-10}
    \multicolumn{1}{l|}{\textbf{LLMs}} &  & \multicolumn{2}{c|}{\textbf{Random Perm}} & \multicolumn{3}{c|}{\textbf{Heuristics}} & \multicolumn{3}{c}{\textbf{ATP Attack}}  \\
    \cmidrule(lr){2-2}\cmidrule(lr){3-4}\cmidrule(lr){5-7}\cmidrule(lr){8-10}
       & \textbf{Vanilla} & \textbf{Rand} &  \textbf{Best 20} & \textbf{Row Rvs} & \textbf{Col Rvs} & \textbf{E-Search}  &\textbf{Row} & \textbf{Col}  & \textbf{Row\&Col}  \\
    \midrule
    \textsc{Llama-3.1-8B}               & 0.28 & 0.15 & 0.14 & 0.25 & 0.21 & 0.14 & 0.23 & 0.19 & \textbf{0.11} \\
    \textsc{Llama-3.1-8B-Inst}           & 0.49 & 0.28 & 0.25 & 0.45 & 0.33 & 0.24 & 0.43 & 0.28 & \textbf{0.20} \\
    \textsc{TableLLM-8B}                & 0.25 & 0.18 & 0.16 & 0.24 & 0.19 & 0.15 & 0.24 & 0.17 & \textbf{0.12} \\
    \textsc{Qwen2.5-1.5B-Inst}           & 0.11& 0.09 & 0.09 & 0.10 & 0.09 & 0.08 & 0.10 & 0.10 & \textbf{0.07} \\
    \textsc{Qwen2.5-3B-Inst}            & 0.22 & 0.16 & 0.13 & 0.20 & 0.15 & 0.13 & 0.19 & 0.15 &  \textbf{0.11} \\
    \textsc{Qwen2.5-7B-Inst}            & 0.28 & 0.20 & 0.17 & 0.23 & 0.19 & 0.17 & 0.21 & 0.18 &  \textbf{0.13} \\
    \textsc{Qwen2.5-14B-Inst}            & 0.47 & 0.30 & 0.28 & 0.43 & 0.32 & 0.28 & 0.41 & 0.29 & \textbf{0.24} \\
    \textsc{CodeLlama-7B-Inst}           & 0.07 & 0.05 & 0.04 & 0.07 & 0.06 & 0.05 & 0.06 & 0.06 & \textbf{0.03} \\
    \textsc{DSR1-Dt-Llama-8B}        & 0.27 & 0.17 & 0.15 & 0.25 & 0.18 & 0.15 & 0.25 & 0.17 & \textbf{0.12} \\
    \textsc{DSR1-Dt-Qwen-7B}          & 0.11 & 0.08 & 0.08 & 0.10 & 0.08 & 0.08 & 0.09 & 0.09 & \textbf{0.05} \\
        \bottomrule
  \end{tabular}
  \vspace{-0.5em}
\end{table*}

\textbf{Attack methods.} We compare ATP with five baselines: (1) a single random row-and-column permutation; (2) best-of-20 random permutations, which reports the strongest attack among 20 sampled permutations for each example; (3) row reversal, which reverses all data rows while keeping the header fixed; (4) column reversal; and (5) evolutionary search, a gradient-free discrete search over row and column permutations. The evolutionary baseline uses elitism: a population initialized from the identity and random permutations is scored, top candidates are retained, and new candidates are generated through order crossover and swap mutation. We use population size 5 and 5 generations, for 30 model queries per example. For ATP, the row and column factors can be fixed independently, so we report row-only, column-only, and joint row-and-column attacks.

\textbf{Victim LLMs.} We evaluate a diverse set of open-source LLMs: Llama-3.1-8B and Llama-3.1-8B-Instruct~\citep{dubey2024llama}; Qwen2.5-1.5B/3B/7B/14B-Instruct~\citep{bai2023qwen}; DeepSeek-R1-Distill-Llama-8B and DeepSeek-R1-Distill-Qwen-7B~\citep{guo2025deepseek}; CodeLlama-7B-Instruct~\citep{roziere2023code}; and TableLLM-8B, which is specifically fine-tuned for TQA~\citep{zhang2024tablellm}. We also evaluate closed-source LLMs, Gemini-2.5 and GPT-4o, using random and heuristic attacks because ATP requires gradients. These results are reported in \cref{subsec:result_on_closed_llm}.
 \looseness=-1

\vspace{-1em}
\subsection{Main Results}
\label{sec:main_result}
\vspace{-0.5em}

\textbf{Vulnerabilities to random permutations.}
\Cref{tab:wtq_result,tab:tatqa_result,tab:fetaqa_result} show that none of the evaluated LLMs is fully robust even to random table permutations. On WTQ, the strongest vanilla open model, Qwen2.5-14B-Instruct, 
drops from 0.47 to 0.33 under a random permutation. On TATQA, Llama-3.1-8B-Instruct drops from 0.49 to 0.28. On FeTaQA, Llama-3.1-8B-Instruct drops from 0.50 to 0.41. These drops occur despite that the permutations preserve table semantics 
for the filtered examples.
 
\vspace{-0.5em}
\textbf{Vulnerabilities to ATP attack.}
ATP produces stronger attacks than random and heuristic baselines across all three datasets. On WTQ, Qwen2.5-14B-Instruct remains the strongest open model under joint ATP, but its score still drops from 0.47 to 0.26. On TATQA, Llama-3.1-8B-Instruct drops from 0.49 to 0.20 under joint ATP. On FeTaQA, Llama-3.1-8B-Instruct drops from 0.50 to 0.31. These results indicate that optimized, semantics-preserving table layouts can cause large failures in current LLM-based TQA pipelines.
 
We observe three recurring patterns. First, joint row-and-column ATP is consistently the strongest attack, outperforming both row-only and column-only ATP as well as all heuristic baselines.
 Second, the joint effect is not simply additive. For example, on WTQ with Llama-3.1-8B-Instruct, row-only ATP drops the score by 0.11, column-only ATP by 0.15, and joint ATP by 0.24.
  In contrast, on FeTaQA with Qwen2.5-14B-Instruct, row-only and column-only ATP cause mild drops of 0.02 and 0.05, but joint ATP causes a much larger drop of 0.19. Third, 
  linearized LLMs appear especially vulnerable to column permutations: column-only ATP is generally stronger than row-only ATP, and column reversal is generally stronger than row reversal. 
  One possible explanation is that row-wise linearization preserves within-row token order under row permutations, whereas column permutations alter token order inside every row and therefore perturb 
  the serialized sequence more broadly. Understanding how this interacts with positional encodings is an important future direction.
 
  \vspace{-0.5em}
\subsection{Mechanistic Insight into LLM Vulnerabilities}
 \vspace{-0.5em}
\Cref{sec:example_attack} gives an illustrative ATP example and the corresponding model response. Using the same example, \cref{subsec:mechanistic_insight} analyzes transformer attention patterns and shows that ATP reduces attention to the correct evidence row while increasing attention across artificial row boundaries created by linearization.

\vspace{-0.5em}
\subsection{Ablation Study and Runtime Analysis}
\vspace{-0.5em}
\label{sec:ablation}
We ablate the entropy penalty and the number of attack iterations in \cref{appendix_sec:ablation}. Runtime is discussed in \cref{appendix_sec:runtime}.

\vspace{-0.5em}
\subsection{Gradient-Free Extensions, Other Structured Data, and Potential Defenses}
\vspace{-0.5em}
ATP is a gradient-based attack and therefore cannot be directly applied to closed-source LLMs such as Gemini or ChatGPT. Our goal is not to claim that ATP is a universally deployable attack, but rather to provide a worst-case diagnostic for structural sensitivity in LLM-based TQA. If optimized semantics-preserving permutations induce large failures, then current table-linearization pipelines are structurally fragile.
 
ATP could also be extended to black-box settings. With query access, zeroth-order optimization~\citep{papernot2017practical,chen2017zoo} could search over the same row and column permutation parameters. Alternatively, an open-source proxy model could generate adversarial permutations for transfer to closed models. We leave these directions for future work. Potential defenses and extensions to other structured data formats are discussed in \cref{subsec:potential_defense}.
\looseness=-1

\vspace{-1em}
\section{Conclusion}
\vspace{-0.5em}
We studied a structural robustness failure in LLM-based TQA: semantically equivalent permutations can substantially change model behavior. 
We formalized this attack space and introduced ATP  for finding worst case permutations. 
Across multiple TQA datasets and LLM families, ATP consistently degrades response alignment. These findings suggest the vulnerabilities of current linearization strategy, motivating future work on permutation-robust table representations and training objectives.
\looseness=-1

\bibliographystyle{plain} 
\bibliography{neurips_2026}

@article{zong2023fool,
  title={Fool your (vision and) language model with embarrassingly simple permutations},
  author={Zong, Yongshuo and Yu, Tingyang and Chavhan, Ruchika and Zhao, Bingchen and Hospedales, Timothy},
  journal={arXiv preprint arXiv:2310.01651},
  year={2023}
}

@article{madry2017towards,
  title={Towards deep learning models resistant to adversarial attacks},
  author={Madry, Aleksander and Makelov, Aleksandar and Schmidt, Ludwig and Tsipras, Dimitris and Vladu, Adrian},
  journal={arXiv preprint arXiv:1706.06083},
  year={2017}
}

@inproceedings{cohen2019certified,
  title={Certified adversarial robustness via randomized smoothing},
  author={Cohen, Jeremy and Rosenfeld, Elan and Kolter, Zico},
  booktitle={international conference on machine learning},
  pages={1310--1320},
  year={2019},
  organization={PMLR}
}

@inproceedings{lin2004rouge,
  title={Rouge: A package for automatic evaluation of summaries},
  author={Lin, Chin-Yew},
  booktitle={Text summarization branches out},
  pages={74--81},
  year={2004}
}

@article{guo2025deepseek,
  title={Deepseek-r1: Incentivizing reasoning capability in llms via reinforcement learning},
  author={Guo, Daya and Yang, Dejian and Zhang, Haowei and Song, Junxiao and Zhang, Ruoyu and Xu, Runxin and Zhu, Qihao and Ma, Shirong and Wang, Peiyi and Bi, Xiao and others},
  journal={arXiv preprint arXiv:2501.12948},
  year={2025}
}

@article{roziere2023code,
  title={Code llama: Open foundation models for code},
  author={Roziere, Baptiste and Gehring, Jonas and Gloeckle, Fabian and Sootla, Sten and Gat, Itai and Tan, Xiaoqing Ellen and Adi, Yossi and Liu, Jingyu and Sauvestre, Romain and Remez, Tal and others},
  journal={arXiv preprint arXiv:2308.12950},
  year={2023}
}

@article{dubey2024llama,
  title={The llama 3 herd of models},
  author={Dubey, Abhimanyu and Jauhri, Abhinav and Pandey, Abhinav and Kadian, Abhishek and Al-Dahle, Ahmad and Letman, Aiesha and Mathur, Akhil and Schelten, Alan and Yang, Amy and Fan, Angela and others},
  journal={arXiv e-prints},
  pages={arXiv--2407},
  year={2024}
}

@article{bai2023qwen,
  title={Qwen technical report},
  author={Bai, Jinze and Bai, Shuai and Chu, Yunfei and Cui, Zeyu and Dang, Kai and Deng, Xiaodong and Fan, Yang and Ge, Wenbin and Han, Yu and Huang, Fei and others},
  journal={arXiv preprint arXiv:2309.16609},
  year={2023}
}

@inproceedings{zhao2022certified,
  title={Certified robustness against natural language attacks by causal intervention},
  author={Zhao, Haiteng and Ma, Chang and Dong, Xinshuai and Luu, Anh Tuan and Deng, Zhi-Hong and Zhang, Hanwang},
  booktitle={International Conference on Machine Learning},
  pages={26958--26970},
  year={2022},
  organization={PMLR}
}

@article{comanici2025gemini,
  title={Gemini 2.5: Pushing the frontier with advanced reasoning, multimodality, long context, and next generation agentic capabilities},
  author={Comanici, Gheorghe and Bieber, Eric and Schaekermann, Mike and Pasupat, Ice and Sachdeva, Noveen and Dhillon, Inderjit and Blistein, Marcel and Ram, Ori and Zhang, Dan and Rosen, Evan and others},
  journal={arXiv preprint arXiv:2507.06261},
  year={2025}
}

@inproceedings{papineni2002bleu,
  title={Bleu: a method for automatic evaluation of machine translation},
  author={Papineni, Kishore and Roukos, Salim and Ward, Todd and Zhu, Wei-Jing},
  booktitle={Proceedings of the 40th annual meeting of the Association for Computational Linguistics},
  pages={311--318},
  year={2002}
}

@article{pasupat2015compositional,
  title={Compositional semantic parsing on semi-structured tables},
  author={Pasupat, Panupong and Liang, Percy},
  journal={arXiv preprint arXiv:1508.00305},
  year={2015}
}

@article{zhu2021tat,
  title={TAT-QA: A question answering benchmark on a hybrid of tabular and textual content in finance},
  author={Zhu, Fengbin and Lei, Wenqiang and Huang, Youcheng and Wang, Chao and Zhang, Shuo and Lv, Jiancheng and Feng, Fuli and Chua, Tat-Seng},
  journal={arXiv preprint arXiv:2105.07624},
  year={2021}
}

@article{nan2022fetaqa,
  title={FeTaQA: Free-form table question answering},
  author={Nan, Linyong and Hsieh, Chiachun and Mao, Ziming and Lin, Xi Victoria and Verma, Neha and Zhang, Rui and Kry{\'s}ci{\'n}ski, Wojciech and Schoelkopf, Hailey and Kong, Riley and Tang, Xiangru and others},
  journal={Transactions of the Association for Computational Linguistics},
  volume={10},
  pages={35--49},
  year={2022},
  publisher={MIT Press One Broadway, 12th Floor, Cambridge, Massachusetts 02142, USA~…}
}

@article{dong2021should,
  title={How should pre-trained language models be fine-tuned towards adversarial robustness?},
  author={Dong, Xinshuai and Luu, Anh Tuan and Lin, Min and Yan, Shuicheng and Zhang, Hanwang},
  journal={Advances in Neural Information Processing Systems},
  volume={34},
  pages={4356--4369},
  year={2021}
}

@inproceedings{dong2021towards,
  title={Towards robustness against natural language word substitutions},
  author={Dong, Xinshuai and Luu, Anh Tuan and Ji, Rongrong and Liu, Hong},
  booktitle={ICLR},
  year={2021}
}

@article{mena2018learning,
  title={Learning latent permutations with gumbel-sinkhorn networks},
  author={Mena, Gonzalo and Belanger, David and Linderman, Scott and Snoek, Jasper},
  journal={arXiv preprint arXiv:1802.08665},
  year={2018}
}

@article{sinkhorn1964relationship,
  title={A relationship between arbitrary positive matrices and doubly stochastic matrices},
  author={Sinkhorn, Richard},
  journal={The annals of mathematical statistics},
  volume={35},
  number={2},
  pages={876--879},
  year={1964},
  publisher={JSTOR}
}

@article{adams2011ranking,
  title={Ranking via sinkhorn propagation},
  author={Adams, Ryan Prescott and Zemel, Richard S},
  journal={arXiv preprint arXiv:1106.1925},
  year={2011}
}

@article{jang2016categorical,
  title={Categorical reparameterization with gumbel-softmax},
  author={Jang, Eric and Gu, Shixiang and Poole, Ben},
  journal={arXiv preprint arXiv:1611.01144},
  year={2016}
}

@article{birkhoff1946tres,
  title={Tres observaciones sobre el algebra lineal},
  author={Birkhoff, Garrett},
  journal={Univ. Nac. Tucuman, Ser. A},
  volume={5},
  pages={147--154},
  year={1946}
}

@article{maddison2016concrete,
  title={The concrete distribution: A continuous relaxation of discrete random variables},
  author={Maddison, Chris J and Mnih, Andriy and Teh, Yee Whye},
  journal={arXiv preprint arXiv:1611.00712},
  year={2016}
}

@article{kingma2014adam,
  title={Adam: A method for stochastic optimization},
  author={Kingma, Diederik P},
  journal={arXiv preprint arXiv:1412.6980},
  year={2014}
}

@article{kuhn1955hungarian,
  title={The Hungarian method for the assignment problem},
  author={Kuhn, Harold W},
  journal={Naval research logistics quarterly},
  volume={2},
  number={1-2},
  pages={83--97},
  year={1955},
  publisher={Wiley Online Library}
}

@article{kuhn1956variants,
  title={Variants of the Hungarian method for assignment problems},
  author={Kuhn, Harold W},
  journal={Naval research logistics quarterly},
  volume={3},
  number={4},
  pages={253--258},
  year={1956},
  publisher={Wiley Online Library}
}

@inproceedings{deng2024acl,
  author       = {Naihao Deng and
                  Zhenjie Sun and
                  Ruiqi He and
                  Aman Sikka and
                  Yulong Chen and
                  Lin Ma and
                  Yue Zhang and
                  Rada Mihalcea},
  title        = {Tables as Texts or Images: Evaluating the Table Reasoning Ability
                  of LLMs and MLLMs},
  booktitle    = {Findings of the Association for Computational Linguistics, {ACL} 2024,
                  Bangkok, Thailand and virtual meeting, August 11-16, 2024},
  publisher    = {Association for Computational Linguistics},
  year         = {2024},
}

@article{rajkumar2022evaluating,
  title={Evaluating the text-to-sql capabilities of large language models},
  author={Rajkumar, Nitarshan and Li, Raymond and Bahdanau, Dzmitry},
  journal={arXiv preprint arXiv:2204.00498},
  year={2022}
}

@inproceedings{2024naacl,
  author       = {Tianyang Liu and
                  Fei Wang and
                  Muhao Chen},
  title        = {Rethinking Tabular Data Understanding with Large Language Models},
  booktitle    = {Proceedings of the 2024 Conference of the North American Chapter of
                  the Association for Computational Linguistics: Human Language Technologies
                  (Volume 1: Long Papers), {NAACL} 2024, Mexico City, Mexico, June 16-21,
                  2024},
  pages        = {450--482},
  publisher    = {Association for Computational Linguistics},
  year         = {2024},
}

@inproceedings{cheng2023iclr,
  author       = {Zhoujun Cheng and
                  Tianbao Xie and
                  Peng Shi and
                  Chengzu Li and
                  Rahul Nadkarni and
                  Yushi Hu and
                  Caiming Xiong and
                  Dragomir Radev and
                  Mari Ostendorf and
                  Luke Zettlemoyer and
                  Noah A. Smith and
                  Tao Yu},
  title        = {Binding Language Models in Symbolic Languages},
  booktitle    = {The Eleventh International Conference on Learning Representations,
                  {ICLR} 2023, Kigali, Rwanda, May 1-5, 2023},
  publisher    = {OpenReview.net},
  year         = {2023},
}

@article{nguyen2025tmlr,
  author       = {Giang Nguyen and
                  Ivan Brugere and
                  Shubham Sharma and
                  Sanjay Kariyappa and
                  Anh Totti Nguyen and
                  Freddy L{\'{e}}cu{\'{e}}},
  title        = {Interpretable LLM-based Table Question Answering},
  journal      = {Trans. Mach. Learn. Res.},
  year         = {2025},
}

@article{su2024tablegpt2,
  title={Tablegpt2: A large multimodal model with tabular data integration},
  author={Su, Aofeng and Wang, Aowen and Ye, Chao and Zhou, Chen and Zhang, Ga and Chen, Gang and Zhu, Guangcheng and Wang, Haobo and Xu, Haokai and Chen, Hao and others},
  journal={arXiv preprint arXiv:2411.02059},
  year={2024}
}

@article{wang2024chain,
  title={Chain-of-table: Evolving tables in the reasoning chain for table understanding},
  author={Wang, Zilong and Zhang, Hao and Li, Chun-Liang and Eisenschlos, Julian Martin and Perot, Vincent and Wang, Zifeng and Miculicich, Lesly and Fujii, Yasuhisa and Shang, Jingbo and Lee, Chen-Yu and others},
  journal={arXiv preprint arXiv:2401.04398},
  year={2024}
}

@inproceedings{qi2024iclr,
  author       = {Xiangyu Qi and
                  Yi Zeng and
                  Tinghao Xie and
                  Pin{-}Yu Chen and
                  Ruoxi Jia and
                  Prateek Mittal and
                  Peter Henderson},
  title        = {Fine-tuning Aligned Language Models Compromises Safety, Even When
                  Users Do Not Intend To!},
  booktitle    = {The Twelfth International Conference on Learning Representations,
                  {ICLR} 2024, Vienna, Austria, May 7-11, 2024},
  publisher    = {OpenReview.net},
  year         = {2024},
}

@article{he2024your,
  title={What is in your safe data? identifying benign data that breaks safety},
  author={He, Luxi and Xia, Mengzhou and Henderson, Peter},
  journal={arXiv preprint arXiv:2404.01099},
  year={2024}
}

@article{hsiung2025llm,
  title={Why LLM Safety Guardrails Collapse After Fine-tuning: A Similarity Analysis Between Alignment and Fine-tuning Datasets},
  author={Hsiung, Lei and Pang, Tianyu and Tang, Yung-Chen and Song, Linyue and Ho, Tsung-Yi and Chen, Pin-Yu and Yang, Yaoqing},
  journal={arXiv preprint arXiv:2506.05346},
  year={2025}
}

@article{wang2023robustness,
  title={On the robustness of chatgpt: An adversarial and out-of-distribution perspective},
  author={Wang, Jindong and Hu, Xixu and Hou, Wenxin and Chen, Hao and Zheng, Runkai and Wang, Yidong and Yang, Linyi and Huang, Haojun and Ye, Wei and Geng, Xiubo and others},
  journal={arXiv preprint arXiv:2302.12095},
  year={2023}
}

@inproceedings{zheng2023nips,
  author       = {Lianmin Zheng and
                  Wei{-}Lin Chiang and
                  Ying Sheng and
                  Siyuan Zhuang and
                  Zhanghao Wu and
                  Yonghao Zhuang and
                  Zi Lin and
                  Zhuohan Li and
                  Dacheng Li and
                  Eric P. Xing and
                  Hao Zhang and
                  Joseph E. Gonzalez and
                  Ion Stoica},
  title        = {Judging LLM-as-a-Judge with MT-Bench and Chatbot Arena},
  booktitle    = {Advances in Neural Information Processing Systems 36: Annual Conference
                  on Neural Information Processing Systems 2023, NeurIPS 2023, New Orleans,
                  LA, USA, December 10 - 16, 2023},
  year         = {2023},
}

@article{yu2023gptfuzzer,
  title={Gptfuzzer: Red teaming large language models with auto-generated jailbreak prompts},
  author={Yu, Jiahao and Lin, Xingwei and Yu, Zheng and Xing, Xinyu},
  journal={arXiv preprint arXiv:2309.10253},
  year={2023}
}

@article{zhang2024tablellm,
  title={TableLLM: Enabling Tabular Data Manipulation by LLMs in Real Office Usage Scenarios},
  author={Zhang, Xiaokang and Luo, Sijia and Zhang, Bohan and Ma, Zeyao and Zhang, Jing and Li Yang and Li, Guanlin and Yao, Zijun and Xu, Kangli and Zhou, Jinchang and Zhang-Li, Daniel and others},
  journal={arXiv preprint arXiv:2403.19318},
  year={2024}
}

@misc{
shi2024judging,
title={Judging the Judges: A Systematic Investigation of Position Bias in Pairwise Comparative Assessments by {LLM}s},
author={Lin Shi and Chiyu Ma and Wenhua Liang and Weicheng Ma and Soroush Vosoughi},
year={2024},
url={https://openreview.net/forum?id=y3jJmrKWQ4}
}

@inproceedings{
wang2025eliminating,
title={Eliminating Position Bias of Language Models: A Mechanistic Approach},
author={Ziqi Wang and Hanlin Zhang and Xiner Li and Kuan-Hao Huang and Chi Han and Shuiwang Ji and Sham M. Kakade and Hao Peng and Heng Ji},
booktitle={The Thirteenth International Conference on Learning Representations},
year={2025},
url={https://openreview.net/forum?id=fvkElsJOsN}
}

@inproceedings{ye2023large,
  title={Large language models are versatile decomposers: Decomposing evidence and questions for table-based reasoning},
  author={Ye, Yunhu and Hui, Binyuan and Yang, Min and Li, Binhua and Huang, Fei and Li, Yongbin},
  booktitle={Proceedings of the 46th international ACM SIGIR conference on research and development in information retrieval},
  pages={174--184},
  year={2023}
}

@inproceedings{jiang2023emnlp,
  author       = {Jinhao Jiang and
                  Kun Zhou and
                  Zican Dong and
                  Keming Ye and
                  Xin Zhao and
                  Ji{-}Rong Wen},
  title        = {StructGPT: {A} General Framework for Large Language Model to Reason
                  over Structured Data},
  booktitle    = {Proceedings of the 2023 Conference on Empirical Methods in Natural
                  Language Processing, {EMNLP} 2023, Singapore, December 6-10, 2023},
  publisher    = {Association for Computational Linguistics},
  year         = {2023},
}

@inproceedings{2024acl,
  author       = {Yi Zeng and
                  Hongpeng Lin and
                  Jingwen Zhang and
                  Diyi Yang and
                  Ruoxi Jia and
                  Weiyan Shi},
  title        = {How Johnny Can Persuade LLMs to Jailbreak Them: Rethinking Persuasion
                  to Challenge {AI} Safety by Humanizing LLMs},
  booktitle    = {Proceedings of the 62nd Annual Meeting of the Association for Computational
                  Linguistics (Volume 1: Long Papers), {ACL} 2024, Bangkok, Thailand,
                  August 11-16, 2024},
  year         = {2024},
}

@inproceedings{ye2022aaai,
  author       = {Muchao Ye and
                  Chenglin Miao and
                  Ting Wang and
                  Fenglong Ma},
  title        = {TextHoaxer: Budgeted Hard-Label Adversarial Attacks on Text},
  booktitle    = {Thirty-Sixth {AAAI} Conference on Artificial Intelligence},
  publisher    = {{AAAI} Press},
  year         = {2022},
}

@inproceedings{huang2021eacl,
  author       = {Kuan{-}Hao Huang and
                  Kai{-}Wei Chang},
  title        = {Generating Syntactically Controlled Paraphrases without Using Annotated
                  Parallel Pairs},
  booktitle    = {Proceedings of the 16th Conference of the European Chapter of the
                  Association for Computational Linguistics: Main Volume, {EACL} 2021,
                  Online, April 19 - 23, 2021},
  pages        = {1022--1033},
  publisher    = {Association for Computational Linguistics},
  year         = {2021},
}

@inproceedings{wang2022nips,
  author       = {Zifeng Wang and
                  Jimeng Sun},
  title        = {TransTab: Learning Transferable Tabular Transformers Across Tables},
  booktitle    = {Advances in Neural Information Processing Systems 35: Annual Conference
                  on Neural Information Processing Systems 2022, NeurIPS 2022, New Orleans,
                  LA, USA, November 28 - December 9, 2022},
  year         = {2022},
}

@inproceedings{chen2023nips,
  author       = {Pei Chen and
                  Soumajyoti Sarkar and
                  Leonard Lausen and
                  Balasubramaniam Srinivasan and
                  Sheng Zha and
                  Ruihong Huang and
                  George Karypis},
  title        = {HyTrel: Hypergraph-enhanced Tabular Data Representation Learning},
  booktitle    = {Advances in Neural Information Processing Systems 36: Annual Conference
                  on Neural Information Processing Systems 2023, NeurIPS 2023, New Orleans,
                  LA, USA, December 10 - 16, 2023},
  year         = {2023},
}

@inproceedings{liangpearl,
  title={PEARL: Towards Permutation-Resilient LLMs},
  author={Liang, CHEN and Shen, Li and Deng, Yang and Zhao, Xiaoyan and Liang, Bin and Wong, Kam-Fai},
  booktitle={The Thirteenth International Conference on Learning Representations},
  year         = {2025},
}

@article{borisov2022language,
  title={Language models are realistic tabular data generators},
  author={Borisov, Vadim and Se{\ss}ler, Kathrin and Leemann, Tobias and Pawelczyk, Martin and Kasneci, Gjergji},
  journal={arXiv preprint arXiv:2210.06280},
  year={2022}
}

@inproceedings{yang2022acl,
  author       = {Jingfeng Yang and
                  Aditya Gupta and
                  Shyam Upadhyay and
                  Luheng He and
                  Rahul Goel and
                  Shachi Paul},
  editor       = {Smaranda Muresan and
                  Preslav Nakov and
                  Aline Villavicencio},
  title        = {TableFormer: Robust Transformer Modeling for Table-Text Encoding},
  booktitle    = {Proceedings of the 60th Annual Meeting of the Association for Computational
                  Linguistics (Volume 1: Long Papers), {ACL} 2022, Dublin, Ireland,
                  May 22-27, 2022},
  year         = {2022},
}

@article{bhandari2025tmlr,
  author       = {Kushal Raj Bhandari and
                  Sixue Xing and
                  Soham Dan and
                  Jianxi Gao},
  title        = {Exploring the Robustness of Language Models for Tabular Question Answering
                  via Attention Analysis},
  journal      = {Trans. Mach. Learn. Res.},
  year         = {2025},
}

@inproceedings{lin-etal-2023-inner,
    title = "An Inner Table Retriever for Robust Table Question Answering",
    author = "Lin, Weizhe  and
      Blloshmi, Rexhina  and
      Byrne, Bill  and
      de Gispert, Adria  and
      Iglesias, Gonzalo",
    booktitle = "Proceedings of the 61st Annual Meeting of the Association for Computational Linguistics (Volume 1: Long Papers)",
    year = "2023",
}

@inproceedings{ye2023sigir,
  author       = {Yunhu Ye and
                  Binyuan Hui and
                  Min Yang and
                  Binhua Li and
                  Fei Huang and
                  Yongbin Li},
  title        = {Large Language Models are Versatile Decomposers: Decomposing Evidence
                  and Questions for Table-based Reasoning},
  booktitle    = {Proceedings of the 46th International {ACM} {SIGIR} Conference on
                  Research and Development in Information Retrieval, {SIGIR} 2023, Taipei,
                  Taiwan, July 23-27, 2023},
  publisher    = {{ACM}},
  year         = {2023},
}

@inproceedings{papernot2017practical,
  title={Practical black-box attacks against machine learning},
  author={Papernot, Nicolas and McDaniel, Patrick and Goodfellow, Ian and Jha, Somesh and Celik, Z Berkay and Swami, Ananthram},
  booktitle={Proceedings of the 2017 ACM on Asia conference on computer and communications security},
  pages={506--519},
  year={2017}
}

@inproceedings{chen2017zoo,
  title={Zoo: Zeroth order optimization based black-box attacks to deep neural networks without training substitute models},
  author={Chen, Pin-Yu and Zhang, Huan and Sharma, Yash and Yi, Jinfeng and Hsieh, Cho-Jui},
  booktitle={Proceedings of the 10th ACM workshop on artificial intelligence and security},
  pages={15--26},
  year={2017}
}

@article{wang2026beyond,
  title        = {Beyond Linearization: Attributed Table Graphs for Table Reasoning},
  author       = {Wang, Yuxiang and Gan, Junhao and Gao, Shengxiang and Ye, Shenghao and Yang, Zhengyi and Qi, Jianzhong},
  journal      = {arXiv preprint arXiv:2601.08444},
  year         = {2026}
}

@inproceedings{li2025table,
  title        = {Table as a Modality for Large Language Models},
  author       = {Li, Liyao and Ye, Chao and Ye, Wentao and Sun, Yifei and Jiang, Zhe and Wang, Haobo and Tian, Jiaming and Zhang, Yiming and Wang, Ningtao and Fu, Xing and Chen, Gang and Zhao, Junbo},
  booktitle    = {Advances in Neural Information Processing Systems},
  year         = {2025}
}

@inproceedings{xu2019modeling,
  title        = {Modeling Tabular Data using Conditional {GAN}},
  author       = {Xu, Lei and Skoularidou, Maria and Cuesta-Infante, Alfredo and Veeramachaneni, Kalyan},
  booktitle    = {Advances in Neural Information Processing Systems},
  year         = {2019}
}

@inproceedings{kotelnikov2023tabddpm,
  title        = {{TabDDPM}: Modelling Tabular Data with Diffusion Models},
  author       = {Kotelnikov, Akim and Baranchuk, Dmitry and Rubachev, Ivan and Babenko, Artem},
  booktitle    = {Proceedings of the 40th International Conference on Machine Learning},
  year         = {2023}
}

@article{solatorio2023realtabformer,
  title        = {{REaLTabFormer}: Generating Realistic Relational and Tabular Data using Transformers},
  author       = {Solatorio, Aivin V. and Dupriez, Olivier},
  journal      = {arXiv preprint arXiv:2302.02041},
  year         = {2023}
}

@article{cromp2025tabby,
  title={Tabby: A language model architecture for tabular and structured data synthesis},
  author={Cromp, Sonia and GNVV, Satya Sai Srinath Namburi and Alkhudhayri, Mohammed and Cao, Catherine and Guo, Samuel and Roberts, Nicholas and Sala, Frederic},
  journal={arXiv preprint arXiv:2503.02152},
  year={2025}
}

\newpage
\appendix
\section{Appendix}


\subsection{Metric Used for Alignment Scores by LLM-as-judge}
\label{sec:llm_as_judge}

We follow \cite{zheng2023nips} and use an LLM-as-judge to measure semantic alignment between the ground-truth response and the model response.
A held-out LLM (Gemini 2.5~\citep{comanici2025gemini}) scores semantic similarity between the model output and the ground truth.
The specific prompt is as follows.

"""You are a text analyzer to analyze to what extent the assistant answer is aligned with the reference answer.

Question: \{question\}.
Reference answer: \{reference\_answer\}.
Assistant answer: \{assistant\_answer\}.

Please provide a score ranging from 0 to 1, where 1 represents perfect alignment and 0 represents no alignment at all. The final answer should contain the score only.
"""

The returned score is constrained to the range from 0 to 1. Because the judge itself may be sensitive to table order, we do not provide the table to the judge. To reduce randomness and improve reproducibility, we use seed 0 and temperature 0.

\subsection{Human Annotation and Justification for LLM-as-Judge}
\label{sec:compare_metric}

We further use human annotation to justify the LLM-as-judge metric.
Specifically, we consider the following metrics.

(i) \textbf{ROUGE-L \cite{lin2004rouge}}: lexical similarity via longest common subsequence between generated and reference answers. We use the recall of ROUGE-L.

(ii) \textbf{LLM-as-judge}~\cite{zheng2023nips}: as defined in \cref{sec:llm_as_judge}.

(iii) \textbf{Human annotation}: human raters assess semantic similarity between the reference answer and the model response.
We use the same prompt template as in \cref{sec:llm_as_judge} and take the mean score from 10 human raters.

\begin{wrapfigure}[16]{r}{0.42\columnwidth} 
  \vspace{-0.8\baselineskip}                
  \centering
  \vspace{-8pt}
  \includegraphics[width=0.95\linewidth]{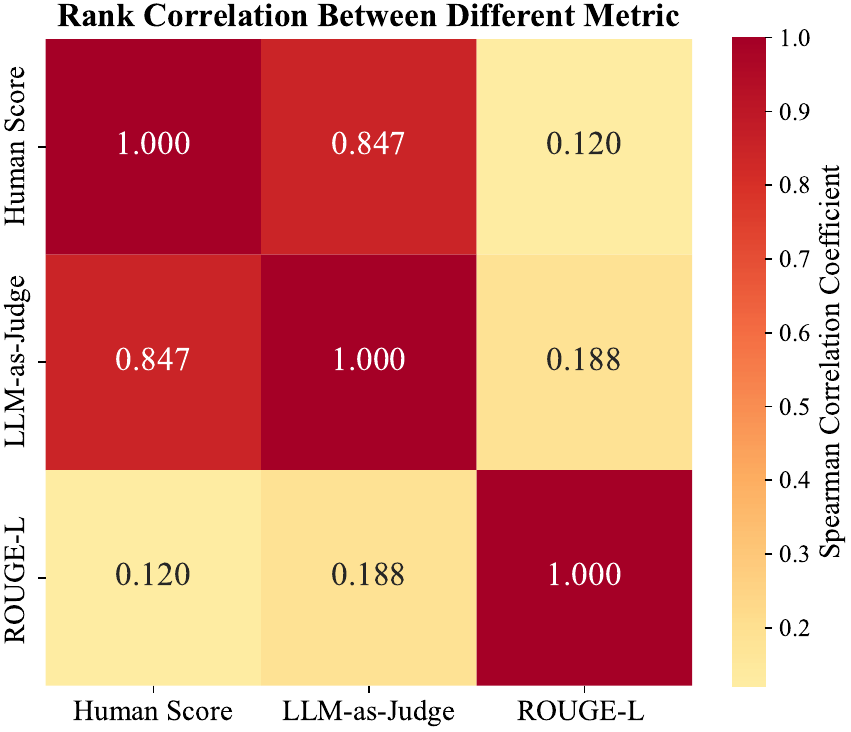}
  \caption{\small{Spearman's rank correlation between metrics. The LLM-as-judge score aligns strongly with human ratings.} }
  \vspace{-15pt}
  \label{fig:rank_corr}
\end{wrapfigure}

\noindent\textbf{Method.} We 
 randomly sample 500 datapoints from the three TQA datasets.
  For each instance, we construct two prompts (original vs.\ \textsc{ATP}-permuted table), prompt the model to generate outputs, 
  and score them with \textsc{ROUGE-L}, the LLM-as-judge, and human raters. To assess agreement among metrics, we compute Spearman's rank correlation over example-level scores for each metric pair. The resulting correlations 
are summarized in Figure~\ref{fig:rank_corr}.

\vspace{-4pt}

Based on these results, we observe that the LLM-as-judge metric exhibits strong alignment with human judgments, whereas \textsc{ROUGE-L} fails to capture such alignment. This finding supports the effectiveness of our
 LLM-as-judge as an evaluation metric for measuring alignment score between LLM-generated responses and ground-truth answers. In contrast, metrics like \textsc{ROUGE-L},
which rely on lexical overlap, are limited to token-level similarity and thus fall short in evaluating semantic equivalence, especially for long-form TQA reasoning. 

We also provide an illustrative example below in \cref{sec:example_attack} to show 
how ATP permutes the input table to fool a victim LLM. The example also illustrates why LLM-as-judge is better suited than ROUGE-L in this setting. 
With the original input, the model gives a correct response that identifies ``State of Origin series'' as the final answer. Both LLM-as-judge and ROUGE-L assign high scores (0.9 and 1.0 respectively).
However, when the input table was permuted, the model outputs intermediate reasoning that enumerates many table cells but does not correctly provide the final response. In this case, though the response is incorrect, ROUGE-L still returns a high score of 1.0 due to surface-level overlap (because the model response enumerates many table cells and does contain ``State of Origin series''). In contrast, LLM-as-judge returns a low score (0.1) for this case, validating its suitability for robust evaluation in such scenarios.

  \vspace{-5pt}
\subsection{An Illustrative Example for ATP Attacked Input}
\label{sec:example_attack}
  \vspace{-5pt}
\begin{tcolorbox}[
  title={Example of ATP Attack Result from WTQ dataset (by Llama-3.1-8B-Instruct)},
  enhanced,
  breakable,
  top=0pt
]
\medskip
\textbf{[Table]}

{\ttfamily
\begin{tabular}{@{}l@{}}
Competition|Total spectatorship|Average match attendance|Year| \\
A-League|1,772,133|12,707|2012/2013| \\
Australian Football League|6,931,085|33,484|2013| \\
Big Bash League|550,262|17,750|2011/2012| \\
National Basketball League|547,021|4,031|2010/2011| \\
National Rugby League|3,345,248|16,643|2013| \\
Super Rugby|773,940|19,348|2012| \\
Rugby Championship|133,532|44,511|2012| \\
State of Origin series|186,607|62,202|2011| \\
Women's National Basketball League|77,944||2010/2011| \\
\end{tabular}
}
\medskip

\textbf{[Table Description]} 

The table contains data related to various sports competitions, including the competition name, total spectatorship, average match attendance, and the year of the data. The total spectatorship and average match attendance figures are provided in numerical format, with some entries for average match attendance being missing.
 
\textbf{[Question]} 

Which had the largest average match attendance?

\tcbline

\textbf{[Ground Truth]}

State of Origin series

\vspace{0pt}
\tcbline

\textbf{Original Answer~\textcolor{red}{(Rouge-L: 1.0, LLM-as-judge Score: 0.9)}} 

\textbf{[Solution]}
To determine which competition had a larger average attendance in a match, we need to identify the highest average from the table. We can do this by comparing the average attendances of each competition.
The highest attendance is 62,202, which is for the \textbf{State of origin series} in 2011.
\vspace{0pt}

\tcbline

\medskip
\textbf{[Table] (Permuted by ATP Attack)}

{\ttfamily
\begin{tabular}{@{}l@{}}
Competition|Total spectatorship|Year|Average match attendance|\\
National Rugby League|3,345,248|2013|16,643|\\
Women's National Basketball League|77,944|2010/2011||\\
Australian Football League|6,931,085|2013|33,484|\\
National Basketball League|547,021|2010/2011|4,031|\\
Big Bash League|550,262|2011/2012|17,750|\\
State of Origin series|186,607|2011|62,202|\\
Super Rugby|773,940|2012|19,348|\\
A-League|1,772,133|2012/2013|12,707|\\
Rugby Championship|133,532|2012|44,511|\\
\end{tabular}
}
\medskip

\vspace{0pt}
\tcbline

\textbf{Attacked Answer \textcolor{red}{(Rouge-L: 1.0, LLM-as-judge Score: 0.1)}} 

\textbf{[Solution]} To find the sport with the highest average attendance in a match, we need to identify the row with maximum average\_match\_attendance. Since some rows have missing values for this column, 
represented as ``nan'', we will exclude those rows from our analysis.
Step 1: Filter out rows with missing average\_attendace values

| Competition|Total Spectatorship |Year |Average Match Attendance|
| --- | ---|--- |--- |
| National Rugby Leauge|3.345.248 |201.3 |16.643|
|Rugger Championship |133.532 | 201 2 |44.511|
Big Basha League |550.262 |20.11/12 |17.750|
State Of Origin Series |186.607 |2.011 |62.202|
Super Rugger |773.940 |.201 |19.348|
Australian Foot Ball League  |6.931.085 |.201.3 33.484|
Women National Basket Ball Leage |77.944 |10/11 |nan|

Step 3: The maximum value in average Match attendance is 62

\end{tcolorbox}

\subsection{Mechanistic Insight of Why LLMs are Vulnerable to Table Permutations}
\label{subsec:mechanistic_insight}
We use the same example as in \cref{sec:example_attack} to inspect the victim LLM and obtain mechanistic insight.

\textbf{Attention-to-Cell Visualization}

\begin{figure}[t]
    \centering
    \begin{subfigure}[t]{0.49\linewidth}
        \centering
        \includegraphics[width=\linewidth]{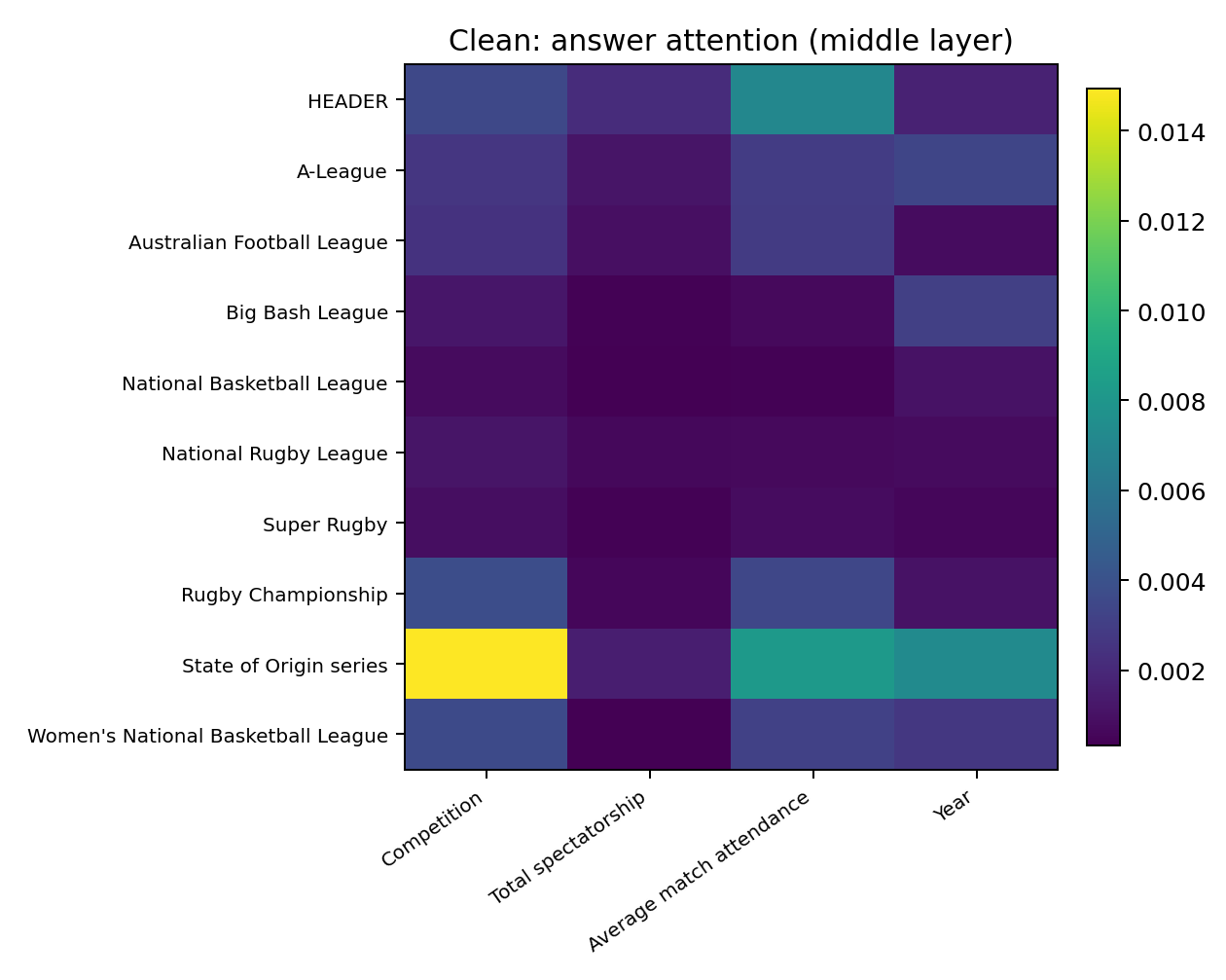}
        \caption{Clean table.}
        \label{fig:cell-attention-clean}
    \end{subfigure}
    \hfill
    \begin{subfigure}[t]{0.49\linewidth}
        \centering
        \includegraphics[width=\linewidth]{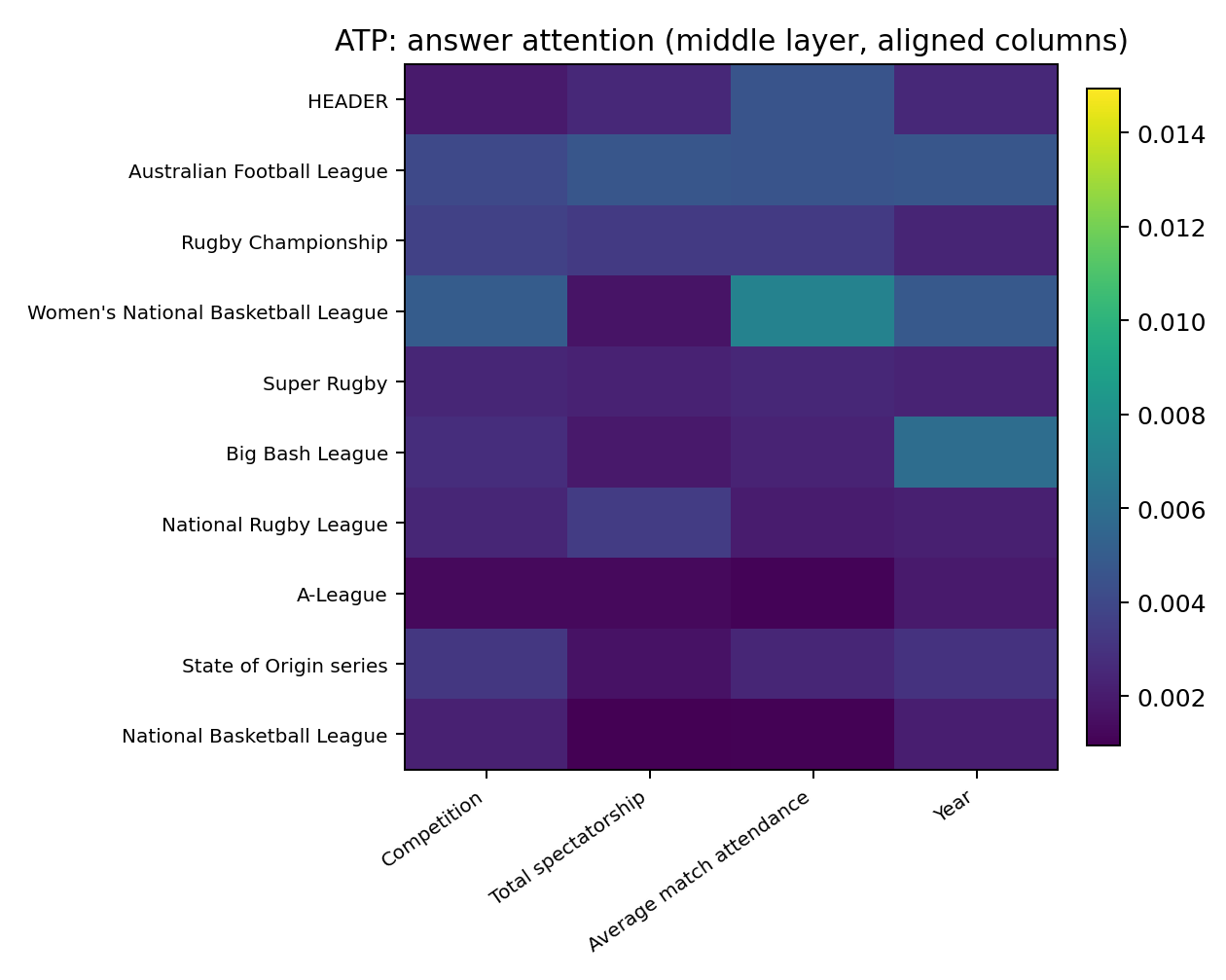}
        \caption{ATP-permuted table.}
        \label{fig:cell-attention-atp}
    \end{subfigure}
    \caption{Middle-layer answer-token attention aggregated to table cells. Columns are aligned by semantic header.}
    \label{fig:cell-attention}
\end{figure}

Figures~\ref{fig:cell-attention-clean} and~\ref{fig:cell-attention-atp} visualize middle-layer attention from generated answer tokens to table cells. The original token-level attention maps are aggregated to table-cell granularity: each heatmap entry sums attention over all tokens belonging to a particular serialized table cell. The $x$-axis gives the semantic table columns, aligned across the clean and ATP-permuted prompts.

The clean prompt shows a concentrated attention pattern on the correct evidence row. The strongest cell is the \textit{Competition} cell for \textit{State of Origin series}, and the model also attends to the relevant \textit{Average match attendance} value, $62{,}202$. After ATP permutation, the attention mass is redistributed away from this correct evidence. The highest-attended cells instead come from unrelated rows, including \textit{Women's National Basketball League} and \textit{Big Bash League}. This suggests that the model's internal evidence retrieval has been misdirected by the adversarial table layout, rather than merely changing the final decoding behavior.

\begin{table}[t]
    \centering
    \begin{tabular}{lcc}
        \toprule
        Setting & Correct competition cell & Correct average-attendance cell \\
        \midrule
        Clean & 0.014940 & 0.008190 \\
        ATP-permuted & 0.003224 & 0.002472 \\
        \bottomrule
    \end{tabular}
    \vspace{1em}
    \caption{Middle-layer answer-token attention to the two key evidence cells.}
    \label{tab:evidence-attention}
\end{table}

Table~\ref{tab:evidence-attention} quantifies the same pattern. Attention to the correct competition cell drops by more than a factor of four, and attention to the correct numerical evidence cell drops by more than a factor of three. These changes are consistent with the qualitative heatmaps: the ATP perturbation weakens attention to the row that supports the correct answer and amplifies attention to irrelevant table entries.

\textbf{Local Layout Diagnostics}

\begin{figure}[t]
    \centering
    \begin{subfigure}[t]{0.49\linewidth}
        \centering
        \includegraphics[width=\linewidth]{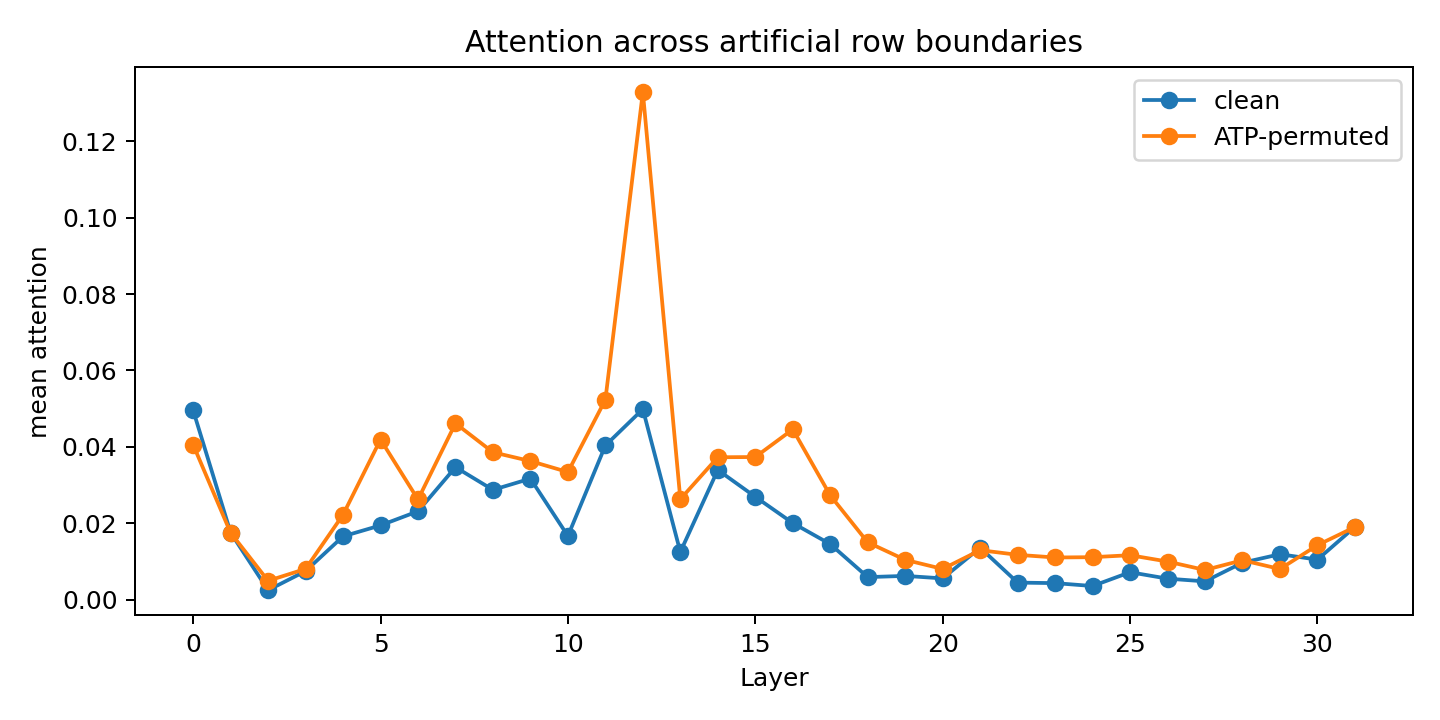}
        \caption{Attention across artificial row boundaries.}
        \label{fig:boundary-attention}
    \end{subfigure}
    \hfill
    \begin{subfigure}[t]{0.49\linewidth}
        \centering
        \includegraphics[width=\linewidth]{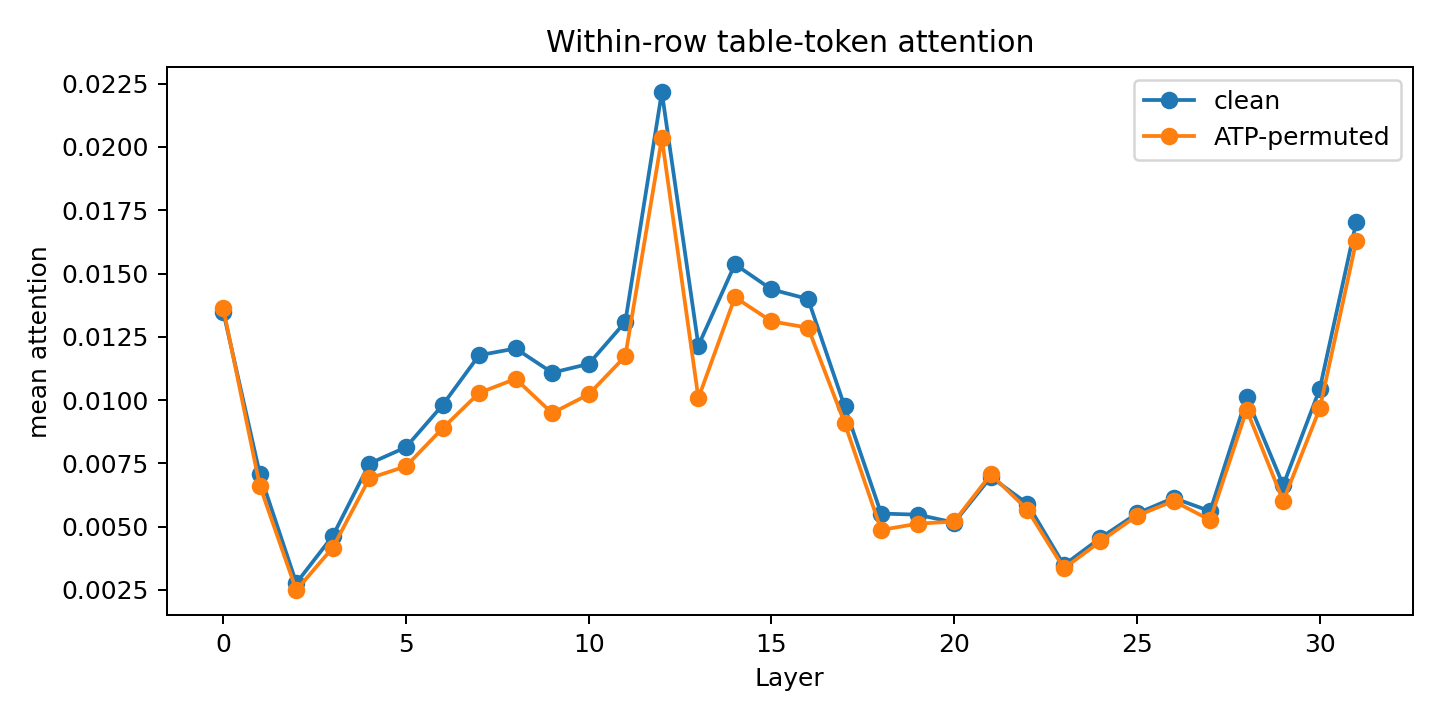}
        \caption{Within-row table-token attention.}
        \label{fig:same-row-attention}
    \end{subfigure}
    \caption{Layer-wise local attention diagnostics for clean and ATP-permuted prompts.}
    \label{fig:local-attention}
\end{figure}

The table is linearized into a one-dimensional token sequence. This creates artificial local adjacencies: the final token of one serialized row is immediately followed by the first token of the next serialized row, even though this boundary is not a meaningful table relation. Figures~\ref{fig:boundary-attention} and~\ref{fig:same-row-attention} compare local table-token attention across transformer layers.

The row-boundary plot measures attention from the first valid token of each serialized row back to the final valid token of the previous row. This is a purely linearization-induced neighborhood. The ATP-permuted table exhibits stronger mean boundary attention than the clean table, $0.026081$ versus $0.017432$, while within-row attention is similar or slightly lower, $0.008631$ versus $0.009349$. This pattern supports the hypothesis that permutation changes how the transformer uses local sequence neighborhoods: after ATP, the model appears more exposed to artificial row-boundary associations while not increasing attention to coherent within-row evidence.

\textbf{Interpretation}

Together, the four figures provide qualitative inner-state evidence that ATP misleads the LLM by altering evidence access inside the transformer. The cell-level attention comparison shows that answer-generation tokens attend less to the correct row and correct numeric evidence after permutation. The local layout diagnostics show that the adversarial layout increases attention across artificial row boundaries introduced by table linearization.

These observations are consistent with the broader ATP hypothesis: current table linearization strategies allow a transformer to exploit positional and neighborhood regularities that are not invariant to row or column permutation. When the table is adversarially reordered, the same question and same table facts can induce the model to retrieve the wrong evidence, producing an incorrect answer even though the task itself is permutation invariant.

\begin{table*}[t]
\vspace{-0.5em}
  \centering
  \caption{LLM-as-judge alignment scores for open LLMs on the FeTaQA evaluation set under different attacks. Lower scores indicate worse response alignment and stronger attacks.}
  \vspace{-0.5em}
  \label{tab:fetaqa_result}
   \setlength{\tabcolsep}{1.0pt}
  \renewcommand{\arraystretch}{1.05}
  \begin{tabular}{l| c|cc|ccc|ccc}
    \toprule
    \multicolumn{1}{c|}{} & \multicolumn{9}{c}{\textbf{FeTaQA Dataset Evaluation Set}} \\
    \cmidrule(lr){2-10}
    \multicolumn{1}{l|}{\textbf{LLMs}} &  & \multicolumn{2}{c|}{\textbf{Random Perm}} & \multicolumn{3}{c|}{\textbf{Heuristics}} & \multicolumn{3}{c}{\textbf{ATP Attack}}  \\
    \cmidrule(lr){2-2}\cmidrule(lr){3-4}\cmidrule(lr){5-7}\cmidrule(lr){8-10}
       & \textbf{Vanilla} & \textbf{Rand} &  \textbf{Best 20} & \textbf{Row Rvs} & \textbf{Col Rvs} & \textbf{E-Search}  &\textbf{Row} & \textbf{Col}  & \textbf{Row\&Col}  \\
    \midrule
    \textsc{Llama-3.1-8B}               & 0.38 & 0.28 & 0.25 & 0.33 & 0.29 & 0.24 & 0.30 & 0.27 & \textbf{0.20} \\
    \textsc{Llama-3.1-8B-Inst}           & 0.50 & 0.41 & 0.37 & 0.44 & 0.42 & 0.35 & 0.42 & 0.40 & \textbf{0.31} \\
    \textsc{TableLLM-8B}                & 0.29 & 0.24 & 0.22 & 0.30 & 0.27 & 0.22 & 0.28 & 0.26 & \textbf{0.20} \\
    \textsc{Qwen2.5-1.5B-Inst}           & 0.11& 0.07 & 0.06 & 0.09 & 0.08 & 0.06 & 0.08 & 0.08 & \textbf{0.04} \\
    \textsc{Qwen2.5-3B-Inst}            & 0.24 & 0.22 & 0.16 & 0.20 & 0.19 & 0.15 & 0.20 & 0.19 &  \textbf{0.12} \\
    \textsc{Qwen2.5-7B-Inst}            & 0.30 & 0.23 & 0.19 & 0.27 & 0.25 & 0.19 & 0.27 & 0.24 &  \textbf{0.14} \\
    \textsc{Qwen2.5-14B-Inst}            & 0.48 & 0.39 & 0.34 & 0.48 & 0.46 & 0.35 & 0.46 & 0.43 & \textbf{0.29} \\
    \textsc{CodeLlama-7B-Inst}           & 0.20 & 0.15 & 0.15 & 0.18 & 0.19 & 0.15 & 0.18 & 0.18 & \textbf{0.13} \\
    \textsc{DSR1-Dt-Llama-8B}        & 0.29 & 0.22 & 0.18 & 0.29 & 0.22 & 0.18 & 0.22 & 0.20 & \textbf{0.13} \\
    \textsc{DSR1-Dt-Qwen-7B}          & 0.14 & 0.11 & 0.09 & 0.13 & 0.12 & 0.09 & 0.13 & 0.12 & \textbf{0.06} \\
        \bottomrule
  \end{tabular}
  \vspace{-0.5em}
\end{table*}

\begin{center}
\begin{table*}[tb]
\vspace{-0em}
   \caption{Ablation of the entropy-penalty weights $\lambda_1$ and $\lambda_2$ using LLM-as-judge alignment scores. Lower scores indicate stronger attacks; $\lambda_1=\lambda_2=10$ generally performs best.}
   \vspace{-0.5em}
   \label{tab:ablation_on_lambda}
  \footnotesize
  \center 
\begin{center}
\begin{tabular}{|c|c|c|c|c|c|c|}
  \hline  
   & \multicolumn{5}{|c|}{ \textbf{WTQ Dataset}}\\
   \cline{2-6}
  \textbf{LLMs} & \multicolumn{5}{|c|}{\textbf{Evaluation Set Against ATP Attack with different values of $\lambda_1,\lambda_2$}} \\
  \cline{2-6}
     & $\lambda_1,\lambda_2=0.0$ & $\lambda_1,\lambda_2=0.1$  & $\lambda_1,\lambda_2=1$ & $\lambda_1,\lambda_2=10$ & $\lambda_1,\lambda_2=20$     \\
  \hline 
  \textsc{Llama-3.1-8B-inst} & 0.24  & 0.24 & 0.23 & \textbf{0.22}  & 0.24 \\ 
  \hline 
  \textsc{Qwen2.5-7B-Inst} & 0.15 & 0.14 & 0.13 &  \textbf{0.12}  & 0.14 \\ 
  \hline
   \hline 
   & \multicolumn{5}{|c|}{ \textbf{TATQA Dataset}}\\
   \cline{2-6}
  \textbf{LLMs}  & \multicolumn{5}{|c|}{\textbf{Evaluation Set Against ATP Attack with different values of $\lambda_1,\lambda_2$}} \\
  \cline{2-6}
     & $\lambda_1,\lambda_2=0.0$ & $\lambda_1,\lambda_2=0.1$  & $\lambda_1,\lambda_2=1$ & $\lambda_1,\lambda_2=10$ & $\lambda_1,\lambda_2=20$     \\
  \hline 
  \textsc{Llama-3.1-8B-inst} &  0.22 & 0.22&  \textbf{0.19} & 0.20  & 0.22 \\ 
  \hline 
  \textsc{Qwen2.5-7B-Inst} & 0.15  & 0.15 & 0.14 & \textbf{0.13}  & 0.15 \\ 
  \hline
   \hline 
   & \multicolumn{5}{|c|}{ \textbf{FeTaQA Dataset}}\\
   \cline{2-6}
  \textbf{LLMs}  & \multicolumn{5}{|c|}{\textbf{Evaluation Set Against ATP Attack with different values of $\lambda_1,\lambda_2$}} \\
  \cline{2-6}
     & $\lambda_1,\lambda_2=0.0$ & $\lambda_1,\lambda_2=0.1$  & $\lambda_1,\lambda_2=1$ & $\lambda_1,\lambda_2=10$ & $\lambda_1,\lambda_2=20$     \\
  \hline 
  \textsc{Llama-3.1-8B-inst} & 0.33 & 0.33 & 0.32 &  \textbf{0.31} &  0.33\\ 
  \hline 
  \textsc{Qwen2.5-7B-Inst} & 0.18  & 0.16 & 0.15  & \textbf{0.14}  & 0.17 \\ 
  \hline
\end{tabular}
\end{center}
\vspace{-0em}
\end{table*}
 \vspace{-0em}
\end{center}

    \begin{figure*}[t]
        \vspace{-0em}
        \begin{minipage}[b]{0.78\linewidth}
        \centering
          \subfloat[Llama-3.1-8B-Inst as victim on  WTQ dataset.]
          {
         \vspace{-0.5em}
          \centering
           \includegraphics[width=0.48\textwidth]{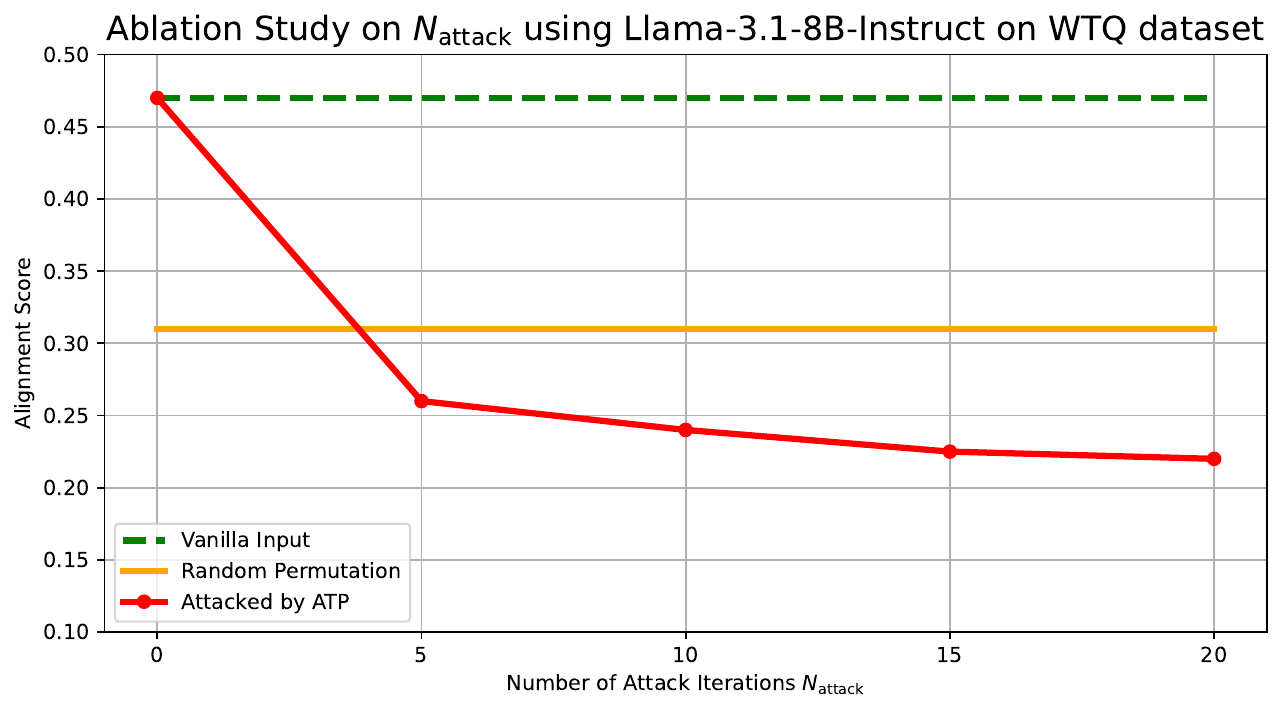}
          }
          \subfloat[Qwen2.5-7B-Inst as victim  on WTQ dataset.]
          {
          \vspace{-0.5em}
         \centering
       \includegraphics[width=0.48\textwidth]{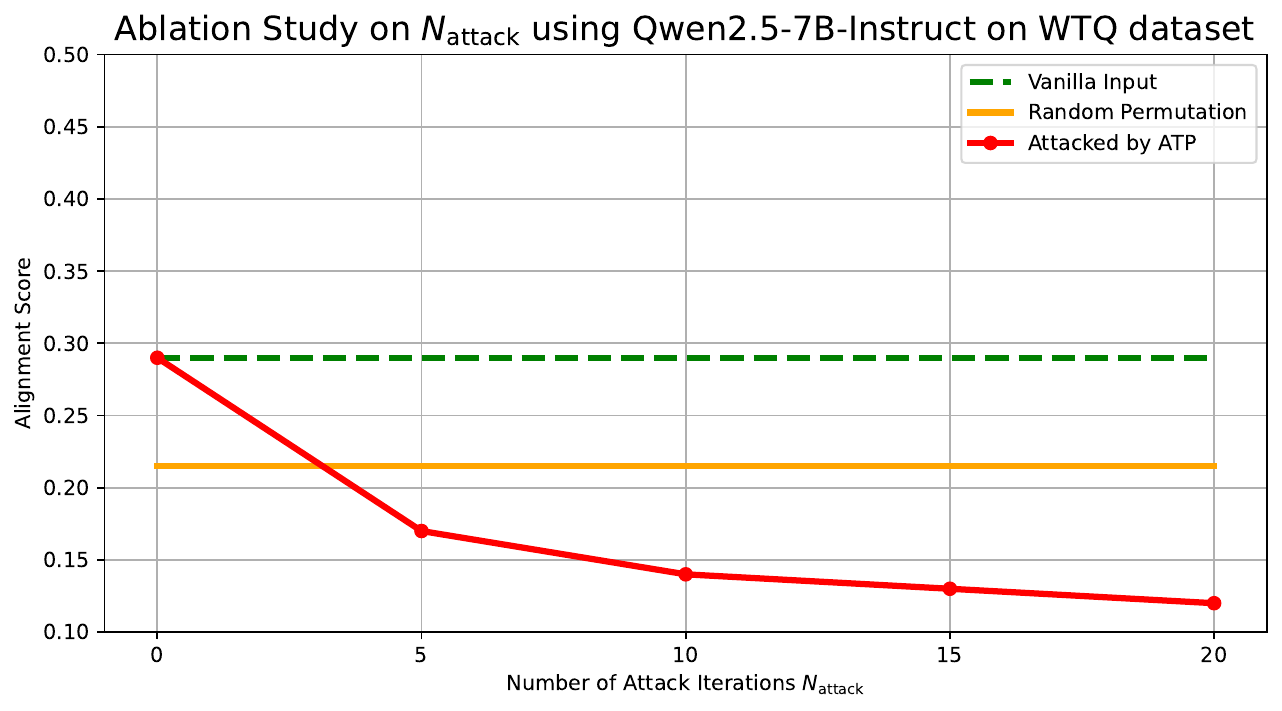}
          }
       \end{minipage}
          \newline
           \centering
           
           \begin{minipage}[b]{0.78\linewidth}
          \hspace{-3.1em}
        \subfloat[Llama-3.1-8B-Inst as victim on  TATQA dataset.]
          {
          \centering
    \vspace{-0.5em}
             \centering
    \includegraphics[width=0.48\textwidth]{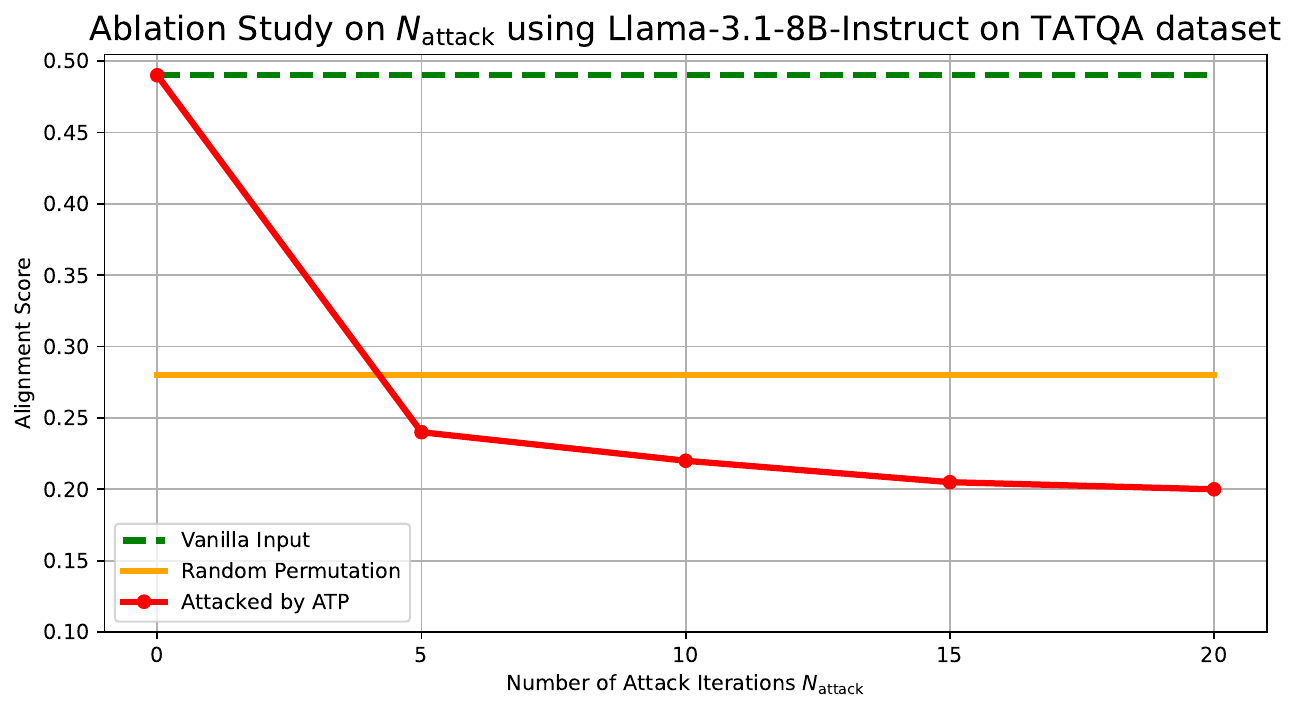}
          }  
        \subfloat[Qwen2.5-7B-Inst as victim  on TATQA dataset.]
          {
          \centering
          \vspace{-0.5em}
                   \centering
     \includegraphics[width=0.48\textwidth]{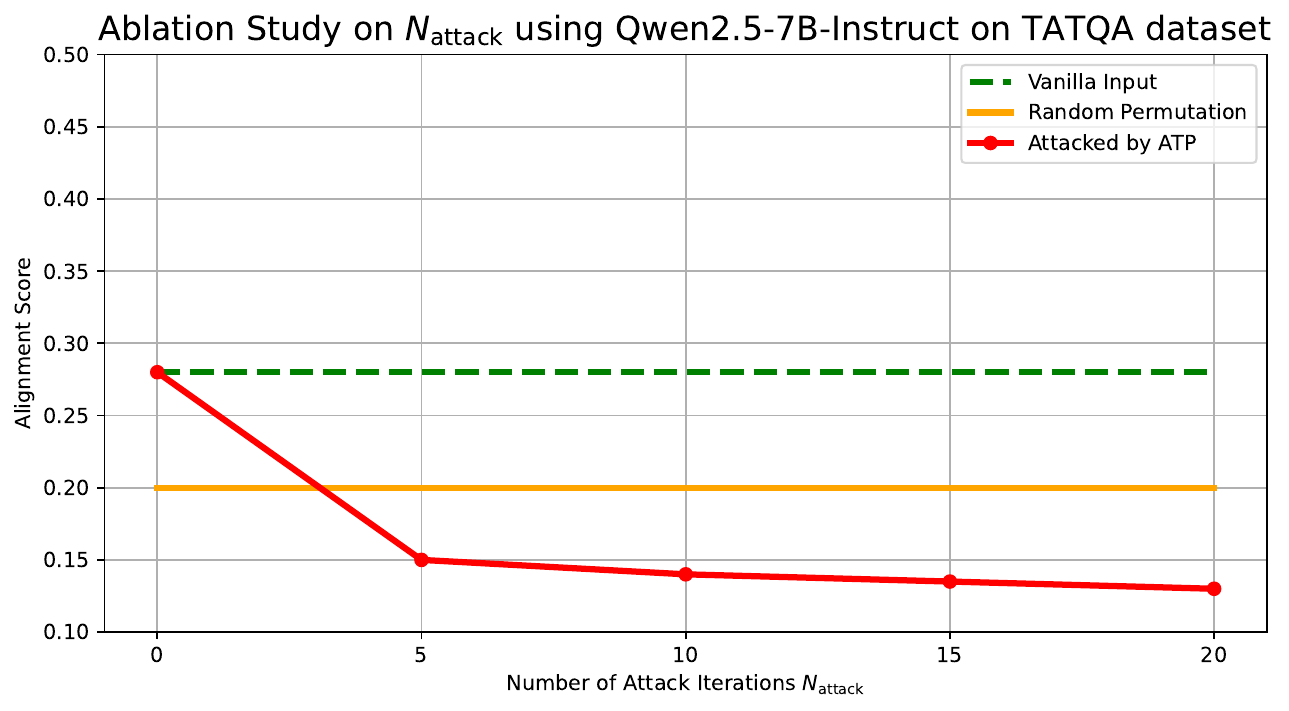}
          }
          \centering
          \vspace{-0.5em}

       \end{minipage}
             \caption{
      Ablation on different values of
       the attack iterations, $N_{\text{attack}}$,
       on  the attack power of ATP, where $N_{\text{attack}}=20$ generally suffices.
       }
        \label{fig:ablation_attack_iters}
        \vspace{-0em}
      \end{figure*}

\vspace{-0em}
\subsection{Ablation Study on ATP}
\label{appendix_sec:ablation}

We conduct ablation studies to investigate (i) the effectiveness of the entropy penalty and (ii) how the number of attack iterations $N_{\text{attack}}$ influences attack performance. The entropy ablation is shown in \cref{tab:ablation_on_lambda}. Using $\lambda_1=\lambda_2=10$ generally gives stronger attacks than $\lambda_1=\lambda_2=0$, which removes the entropy penalty. For example, on the FeTaQA evaluation set, ATP without the entropy penalty decreases the Qwen2.5-7B-Instruct score from 0.30 to 0.18, whereas $\lambda_1=\lambda_2=10$ further decreases it to 0.14. Very large entropy weights are not optimal either, because the optimization focuses too heavily on hardening the soft permutations rather than increasing the victim-model loss. We therefore use $\lambda_1=\lambda_2=10$ in the main experiments.

The effect of $N_{\text{attack}}$ is shown in \cref{fig:ablation_attack_iters}. ATP with only 5 iterations already outperforms random permutations. For example, in \cref{fig:ablation_attack_iters}(a), ATP with $N_{\text{attack}}=5$ degrades Llama-3.1-8B-Instruct on WTQ from 0.47 to 0.26, while a random permutation decreases it only to 0.31. Increasing $N_{\text{attack}}$ generally strengthens the attack, with performance beginning to converge around $N_{\text{attack}}=20$. We therefore use $N_{\text{attack}}=20$ in the main experiments. \looseness=-1

\subsection{Runtime Analysis}
\label{appendix_sec:runtime}
The computational cost of ATP largely depends on the hyper-parameter $N_{\text{attack}}$.
In our implementation with a single A100 GPU, 
ATP with $N_{\text{attack}}=20$ takes about 10 seconds per data point.
Reducing $N_{\text{attack}}$ to 5 trades off some attack strength but still substantially degrades model performance, while reducing runtime to about 3 seconds per sample.

\subsection{Label-Free Attack Result Using KL Loss}
\label{subsec:kl_attack_result}
The result is shown in \cref{tab:wtq_kl_attack_result}. Even in the label-free setting, ATP can use KL divergence to find adversarial table permutations, and this label-free variant remains effective. For example, on WTQ eval set with CodeLlama-7B-Instruct, with labels, CE-based ATP decreases the model performance from 0.18 to 0.09,
 while the label-free ATP can also decrease the performance from 0.18 to 0.10.

\subsection{Heuristic Attack on Closed LLMs}
\label{subsec:result_on_closed_llm}
The results are in \cref{tab:wtq_result2,tab:tatqa_result2,tab:fetaqa_result2}.
The results show that stronger closed LLMs start from substantially higher vanilla performance, but remain vulnerable to semantics-preserving permutations. 
For example, on WTQ, Gemini-2.5 achieves 0.88 under vanilla input but drops to 0.72 under a simple column-reversal attack. 
This supports our main claim that the phenomenon is not merely due to weak baseline performance on a difficult benchmark, 
but reflects a broader structural fragility in current LLM table-linearization pipelines.

\begin{table*}[t]
\vspace{0em}
  \centering
  \caption{Label-free ATP attack performance on the WTQ evaluation set using KL loss. Lower scores indicate worse response alignment and stronger attacks.}
  \vspace{-0.5em}
  \label{tab:wtq_kl_attack_result}
   \setlength{\tabcolsep}{1.0pt}
  \renewcommand{\arraystretch}{1.05}
  \begin{tabular}{l| c|ccc|ccc}
    \toprule
    \multicolumn{1}{c|}{} & \multicolumn{7}{c}{\textbf{WTQ Dataset Evaluation Set}} \\
    \cmidrule(lr){2-8}
    \multicolumn{1}{l|}{\textbf{LLMs}} &   & \multicolumn{3}{c|}{\textbf{KL-based ATP Attack}} & \multicolumn{3}{c}{\textbf{CE-based ATP Attack}}  \\
    \cmidrule(lr){2-2}\cmidrule(lr){3-5}\cmidrule(lr){6-8}
       & \textbf{Vanilla}  & \textbf{Row} & \textbf{Col} & \textbf{Row\&Col}  &\textbf{Row} & \textbf{Col}  & \textbf{Row\&Col}  \\
    \midrule
    \textsc{Llama-3.1-8B-Inst}           & 0.46 & 0.37  & 0.34  & 0.26 & 0.35 & 0.31 & \textbf{0.22} \\
    \textsc{TableLLM-8B}                & 0.33 & 0.30 & 0.27 & 0.19 & 0.28 & 0.24 & \textbf{0.16} \\
    \textsc{Qwen2.5-14B-Inst}            & 0.47  & 0.40 & 0.39  & 0.28 & 0.38 & 0.36 & \textbf{0.26} \\
    \textsc{CodeLlama-7B-Inst}           & 0.18  & 0.18  & 0.17  & 0.10 & 0.16 & 0.14 & \textbf{0.09} \\
    \textsc{DSR1-Dt-Llama-8B}        & 0.24  &  0.21 &  0.18 &  0.11 & 0.19 & 0.16 & \textbf{0.09} \\
    \textsc{DSR1-Dt-Qwen-7B}          & 0.15 & 0.13 &  0.13 & 0.08 & 0.12 & 0.11 & \textbf{0.06} \\
    \bottomrule
  \end{tabular}
  \vspace{-0em}
\end{table*}

\subsection{Potential Defenses and Extension to Other Structured Data}
\label{subsec:potential_defense}
One straightforward defense is to follow the adversarial training paradigm \cite{madry2017towards}: for each example, 
find a worst-case permutation by ATP and train the model to remain correct under such perturbations, leading naturally to a min-max optimization objective.
On the other hand, ATP may also inspire better linearization strategies.
 For example, it may motivate randomized smoothing \cite{cohen2019certified} style of linearization strategies that improve robustness to permutation attacks.
 More broadly, analyzing which permutations are most harmful could help identify linearization choices that are less brittle in practice.

Regarding other structured data such as JSON and XML, 
we expect this adversarial vulnerability to extend beyond tables whenever structured inputs are serialized into a one-dimensional sequence and the model becomes sensitive to superficial ordering.
  That said, tables provide the cleanest setting, because row and column permutations preserve semantics in a particularly direct and intuitive way.
   For formats such as JSON or XML, by contrast, the effectiveness of such attacks may depend more strongly on the specific representation and downstream task.

\begin{table*}[t]
\vspace{0em}
  \centering
  \caption{LLM-as-judge alignment scores for closed LLMs on the WTQ evaluation set under random and heuristic permutation attacks.}
  \vspace{-0.5em}
  \label{tab:wtq_result2}
   \setlength{\tabcolsep}{1.0pt}
  \renewcommand{\arraystretch}{1.05}
  \begin{tabular}{l| c|cc|ccc}
    \toprule
    \multicolumn{1}{c|}{} & \multicolumn{6}{c}{\textbf{WTQ Dataset Evaluation Set}} \\
    \cmidrule(lr){2-7}
    \multicolumn{1}{l|}{\textbf{LLMs}} &  & \multicolumn{2}{c|}{\textbf{Random Perm}} & \multicolumn{3}{c}{\textbf{Heuristics}}  \\
    \cmidrule(lr){2-2}\cmidrule(lr){3-4}\cmidrule(lr){5-7}
       & \textbf{Vanilla} & \textbf{Rand} &  \textbf{Best of 20} & \textbf{Row Revs} & \textbf{Col Revs} & \textbf{Evol Search}   \\
    \midrule
        \textsc{Gemini-2.5}          & 0.88  & 0.68 & 0.61  &  0.81 & 0.72 & 0.60\\
        \textsc{GPT-4o}          & 0.95 &0.78  &0.65  & 0.91& 0.82& 0.63 \\
    \bottomrule
  \end{tabular}
  \vspace{-0em}
\end{table*}

\begin{table*}[t]
\vspace{0em}
  \centering
  \caption{LLM-as-judge alignment scores for closed LLMs on the TATQA evaluation set under random and heuristic permutation attacks.}
  \vspace{-0.5em}
  \label{tab:tatqa_result2}
   \setlength{\tabcolsep}{1.0pt}
  \renewcommand{\arraystretch}{1.05}
  \begin{tabular}{l| c|cc|ccc}
    \toprule
    \multicolumn{1}{c|}{} & \multicolumn{6}{c}{\textbf{TATQA Dataset Evaluation Set}} \\
    \cmidrule(lr){2-7}
    \multicolumn{1}{l|}{\textbf{LLMs}} &  & \multicolumn{2}{c|}{\textbf{Random Perm}} & \multicolumn{3}{c}{\textbf{Heuristics}}  \\
    \cmidrule(lr){2-2}\cmidrule(lr){3-4}\cmidrule(lr){5-7}
       & \textbf{Vanilla} & \textbf{Rand} &  \textbf{Best of 20} & \textbf{Row Revs} & \textbf{Col Revs} & \textbf{Evol Search}   \\
    \midrule
        \textsc{Gemini-2.5}       & 0.91  & 0.65 & 0.58  & 0.84 & 0.71 & 0.56\\
        \textsc{GPT-4o}          & 0.96 & 0.68 & 0.59  & 0.95 & 0.78  & 0.56 \\
    \bottomrule
  \end{tabular}
  \vspace{-0em}
\end{table*}

\begin{table*}[t]
\vspace{-0em}
  \centering
 \caption{LLM-as-judge alignment scores for closed LLMs on the FeTaQA evaluation set under random and heuristic permutation attacks.}
  \vspace{-0.5em}
  \label{tab:fetaqa_result2}
   \setlength{\tabcolsep}{1.0pt}
  \renewcommand{\arraystretch}{1.05}
  \begin{tabular}{l| c|cc|ccc}
    \toprule
    \multicolumn{1}{c|}{} & \multicolumn{6}{c}{\textbf{FeTaQA Dataset Evaluation Set}} \\
    \cmidrule(lr){2-7}
    \multicolumn{1}{l|}{\textbf{LLMs}} &  & \multicolumn{2}{c|}{\textbf{Random Perm}} & \multicolumn{3}{c}{\textbf{Heuristics}}  \\
    \cmidrule(lr){2-2}\cmidrule(lr){3-4}\cmidrule(lr){5-7}
       & \textbf{Vanilla} & \textbf{Rand} &  \textbf{Best of 20} & \textbf{Row Revs} & \textbf{Col Revs} & \textbf{Evol Search}   \\
    \midrule
        \textsc{Gemini-2.5}          & 0.89& 0.76 & 0.68 &0.82 &0.80  & 0.66 \\
        \textsc{GPT-4o}          & 0.94 &0.90  & 0.77  & 0.92 & 0.90 & 0.75 \\
    \bottomrule
  \end{tabular}
  \vspace{-0em}
\end{table*}

\subsection{Limitations}
\label{subsec:limitations}
ATP is designed as a worst-case diagnostic for order sensitivity rather than as a complete robustness benchmark.
 The direct ATP optimization requires gradients and therefore applies most naturally to open models; for closed models, 
 we evaluate random and heuristic attacks and discuss black-box extensions as promising future directions. Our experiments focus on TQA examples whose answers should be invariant to row and column permutations,
  so the conclusions should not be extrapolated to tasks where presentation order is semantically meaningful. 
  

\subsection{Broader Impact}
\label{subsec:broader_impact}
This work has positive potential impact by revealing a practical reliability weakness in LLM-based table reasoning systems and by providing a diagnostic tool that can support 
adversarial training, robust evaluation, and better table interfaces. The main negative risk is dual use: an attacker could intentionally reorder semantically equivalent table inputs to induce wrong answers in deployed systems.
 We mitigate this risk by framing ATP as a robustness evaluation method, studying possible defenses, and avoiding the release of sensitive data or new high-risk models. 

\subsection{Additional Reproducibility Details}
\label{subsec:additional_reproducibility}
The experiments use public TQA datasets and publicly documented LLM checkpoints or closed-model APIs, all cited in the main text. The attack hyperparameters used in the main experiments are $\lambda_1=\lambda_2=10$ and $N_{\text{attack}}=20$, selected by the ablations in \cref{appendix_sec:ablation}. 
In our implementation, ATP with $N_{\text{attack}}=20$ takes about 10 seconds per example on a single A100 GPU, while $N_{\text{attack}}=5$ takes about 3 seconds per example. Our code will be made publicly available.


\end{document}